\documentclass[10pt,twocolumn,letterpaper]{article}
\usepackage[pagenumbers]{cvpr} 
\usepackage{times}

\usepackage{color}
\usepackage[dvipsnames]{xcolor}
\usepackage{epsfig}
\usepackage{graphicx}

\usepackage{adjustbox}
\usepackage{array}
\usepackage{booktabs}
\usepackage{colortbl}
\usepackage{float,wrapfig}
\usepackage{hhline}
\usepackage{multirow}
\usepackage{caption}
\usepackage{amsmath,amsfonts,amsthm,amssymb}
\usepackage{bm}
\usepackage{nicefrac}
\usepackage{microtype}
\usepackage{arydshln}

\usepackage{changepage}
\usepackage{extramarks}
\usepackage{fancyhdr}
\usepackage{lastpage}
\usepackage{setspace}
\usepackage{soul}
\usepackage{xspace}
\usepackage{dsfont}

\usepackage{algorithm}
\usepackage{algpseudocode}
\usepackage{enumerate}
\usepackage{pifont}

\usepackage{url}
\usepackage[pagebackref=false,breaklinks=true,colorlinks,bookmarks=false]{hyperref}
\usepackage{paralist}

\newcommand{\G}{G\xspace}
\newcommand{\latent}{\rvz\xspace}
\newcommand{\real}{\rvx\xspace}
\newcommand{\realdist}{p_{\text{data}}(\rvx)\xspace}
\newcommand{\realfeat}{\rvf\xspace}
\newcommand{\realLR}{\overline{\rvx}\xspace}
\newcommand{\HLR}{\overline{H}\xspace}
\newcommand{\WLR}{\overline{W}\xspace}

\newcommand{\fakefeat}{\hat{\rvf}\xspace}

\newcommand{\quantizeRef}{Q \xspace}
\newcommand{\quantizeTest}{\widehat{\quantizeRef} \xspace}

\newcommand{\resize}{\psi \xspace}
\newcommand{\resizedata}{\resize_{\text{data}}\xspace}
\newcommand{\resizeFID}{\resize_{\text{FID}}\xspace}
\newcommand{\resizerealFID}{\resize_{\text{FID}}\xspace}
\newcommand{\resizefakeFID}{\widehat{\resize}_{\text{FID}}\xspace}
\newcommand{\F}{\mathcal{F}\xspace}
\newcommand{\reponame}{clean-fid\xspace}

\newcommand{\IconCheck}{\includegraphics[height=2.8mm, trim=0 2.5mm 0 0mm] {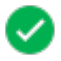}}

\newcommand{\IconCross}{\includegraphics[height=2.8mm, trim=0 2.5mm 0 0mm]{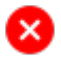}}

\newcommand{\myparagraph}[1]{\vspace{-6pt}\paragraph{#1}}

\newcommand{\ignorethis}[1]{}

\makeatletter
\DeclareRobustCommand\onedot{\futurelet\@let@token\@onedot}
\def\@onedot{\ifx\@let@token.\else.\null\fi\xspace}

\makeatother

\newcommand{\updates}[1]{{\color{black}{#1}}}

\makeatletter
\newcommand\footnoteref[1]{\protected@xdef\@thefnmark{\ref{#1}}\@footnotemark}
\makeatother

\definecolor{black}{rgb}{0.0,0.0,0.0}
\definecolor{mydarkblue}{rgb}{0.0,0.0,.7}
\definecolor{mydarkred}{rgb}{0.8,0.02,0.02}
\definecolor{mydarkorange}{rgb}{0.40,0.2,0.02}
\definecolor{mypurple}{RGB}{111,0,255}
\definecolor{myred}{rgb}{1.0,0.0,0.0}
\definecolor{mygold}{rgb}{0.75,0.6,0.12}
\definecolor{mydarkgray}{rgb}{0.66, 0.66, 0.66}
\definecolor{mygray}{gray}{0.9}
\definecolor{keynotegreen}{rgb}{0.04,0.52,0}
\definecolor{keynoteyellow}{rgb}{1,0.68,0}
\definecolor{fig2_a}{rgb}{237,125,49}
\definecolor{fig2_b}{rgb}{68,114,196}
\definecolor{fig2_c}{rgb}{255,192,0}

\newcommand{\fid}{Fr\'echet Inception Distance\xspace}
\newcommand{\reffig}[1]{Figure~\ref{fig:#1}}
\newcommand{\refsec}[1]{Section~\ref{sec:#1}}

\newcommand{\reftbl}[1]{Table~\ref{tab:#1}}

\newcommand{\lblfig}[1]{\label{fig:#1}}
\newcommand{\lblsec}[1]{\label{sec:#1}}
\newcommand{\lbleq}[1]{\label{eq:#1}}
\newcommand{\lbltbl}[1]{\label{tab:#1}}

\usepackage{etoolbox}
\newcommand{\algorithmicdoinparallel}{\textbf{do in parallel}}
\makeatletter
\AtBeginEnvironment{algorithmic}{%
  \newcommand{\FORP}[2][default]{\ALC@it\algorithmicfor\ #2\ %
    \algorithmicdoinparallel\ALC@com{#1}\begin{ALC@for}}%
}
\makeatother

\newcommand{\abs}[1]{\lvert#1\rvert}

\usepackage{mdframed}

\newcommand{\be}{\begin{equation}}
\newcommand{\ee}{\end{equation}}
\definecolor{Gray}{gray}{0.85}
\definecolor{LightCyan}{rgb}{0.88,1,1}

\makeatletter
\usepackage{xspace} 
\def\@onedot{\ifx\@let@token.\else.\null\fi\xspace}
\DeclareRobustCommand\onedot{\futurelet\@let@token\@onedot}

\definecolor{blue1}{RGB}{0,128,255}
\definecolor{blue3}{RGB}{0,0,128}
\definecolor{darkpastelgreen}{rgb}{0.01, 0.75, 0.24}
\definecolor{cerulean}{rgb}{0.0, 0.48, 0.65}

\def\1{\bm{1}}

\def\rvf{{\mathbf{f}}}

\def\rvx{{\mathbf{x}}}

\def\rvz{{\mathbf{z}}}

\DeclareMathAlphabet{\mathsfit}{\encodingdefault}{\sfdefault}{m}{sl}
\SetMathAlphabet{\mathsfit}{bold}{\encodingdefault}{\sfdefault}{bx}{n}

\def\confName{CVPR}
\def\confYear{2022}

\begin{document}

\title{On Aliased Resizing and Surprising Subtleties in GAN Evaluation}
\author{Gaurav Parmar$^1$ \qquad Richard Zhang$^2$ \qquad Jun-Yan Zhu$^1$ \\
$^1$Carnegie Mellon University \qquad $^2$Adobe Research \qquad\\}

\captionsetup[figure]{font=small}
\captionsetup[table]{font=small}
\maketitle

\begin{abstract}

Metrics for evaluating generative models aim to measure the discrepancy between real and generated images. The often-used \fid (FID) metric, for example, extracts ``high-level'' features using a deep network from the two sets. However, we find that the differences in ``low-level'' preprocessing, specifically image resizing and compression, can induce large variations and have unforeseen consequences. For instance, when resizing an image, e.g., with a bilinear or bicubic kernel, signal processing principles mandate adjusting prefilter width depending on the downsampling factor, to antialias to the appropriate bandwidth. 
However, commonly-used implementations use a fixed-width prefilter, resulting in aliasing artifacts.
Such aliasing leads to corruptions in the feature extraction downstream. Next, lossy compression, such as JPEG, is commonly used to reduce the file size of an image. Although designed to minimally degrade the perceptual quality of an image, the operation also produces variations downstream. Furthermore, we show that if compression is used on real training images, FID can actually improve if the generated images are also subsequently compressed. This paper shows that choices in low-level image processing have been an underappreciated aspect of generative modeling. We identify and characterize variations in generative modeling development pipelines, provide recommendations based on signal processing principles, and release a reference implementation to facilitate future comparisons. 

\end{abstract}

\begin{figure}[t]
    \centering
    \includegraphics[width=0.95\linewidth]{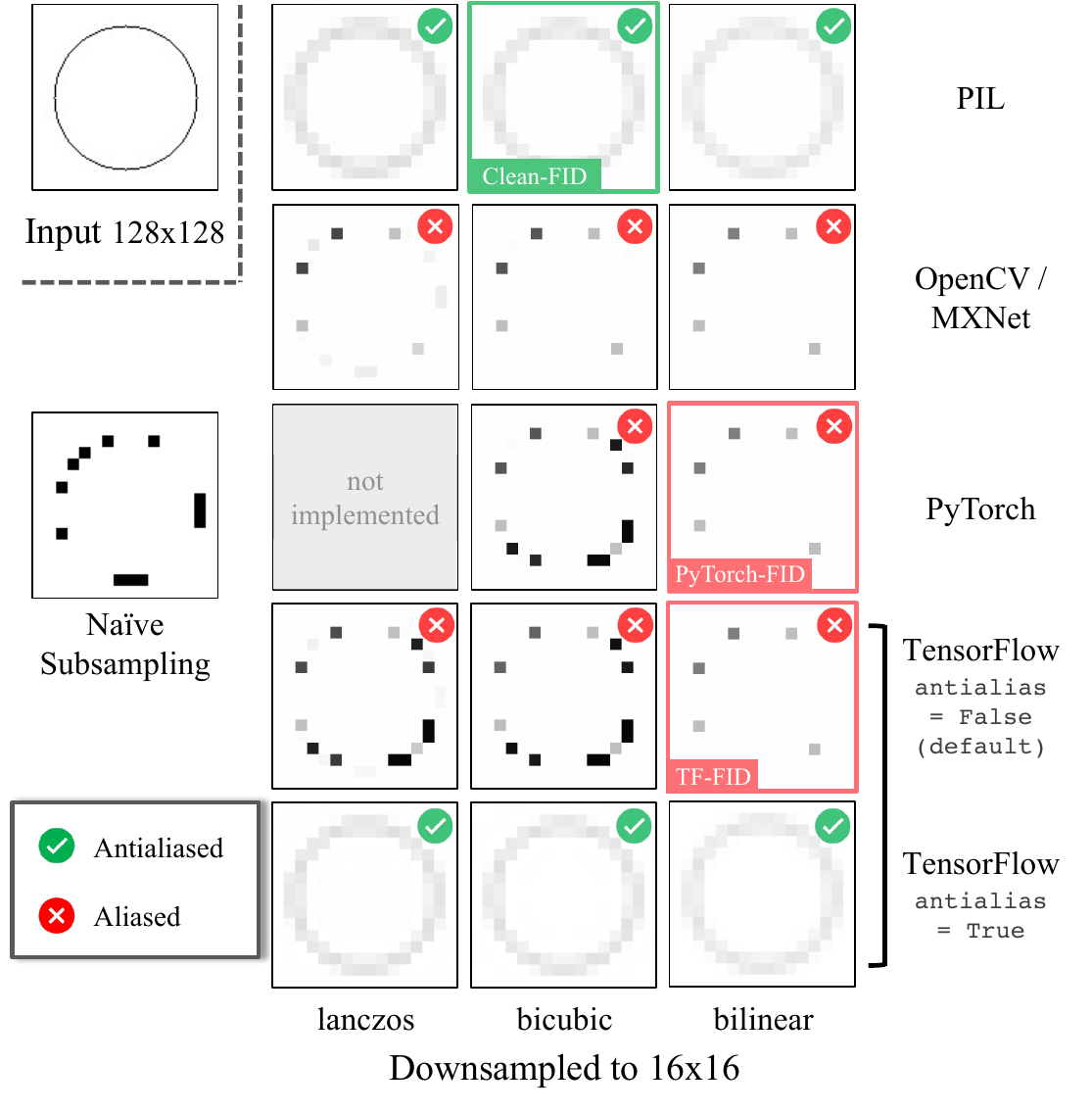}
    \vspace{-10pt}
    \caption{
    \textbf{Downsampling a circle.} We resize an input image (left) by a factor of 8, using different image processing libraries. 
    The Lanczos, bicubic, and bilinear implementations by PIL (top row) adjust the antialiasing filter width by the downsampling factor (marked as \IconCheck). Other implementations (including those used for PyTorch-FID and TensorFlow-FID) use fixed filter widths, introducing aliasing artifacts (marked as \IconCross) and resembling naive nearest subsampling.
    Aliasing artifacts induce inconsistencies in the calculation of downstream metrics such as \fid~\cite{heusel2017gans}, KID~\cite{binkowski2018demystifying}, IS~\cite{salimans2016improved}, and PPL~\cite{karras2019style}. Note that \texttt{antialias} flag is available in TensorFlow 2, but is set to \texttt{False} (default value) for the FID calculation.
    }
    \lblfig{teaser}
    \vspace{-5pt}
\end{figure}
\begin{figure}[t]
    \centering
    \includegraphics[width=0.99\linewidth]{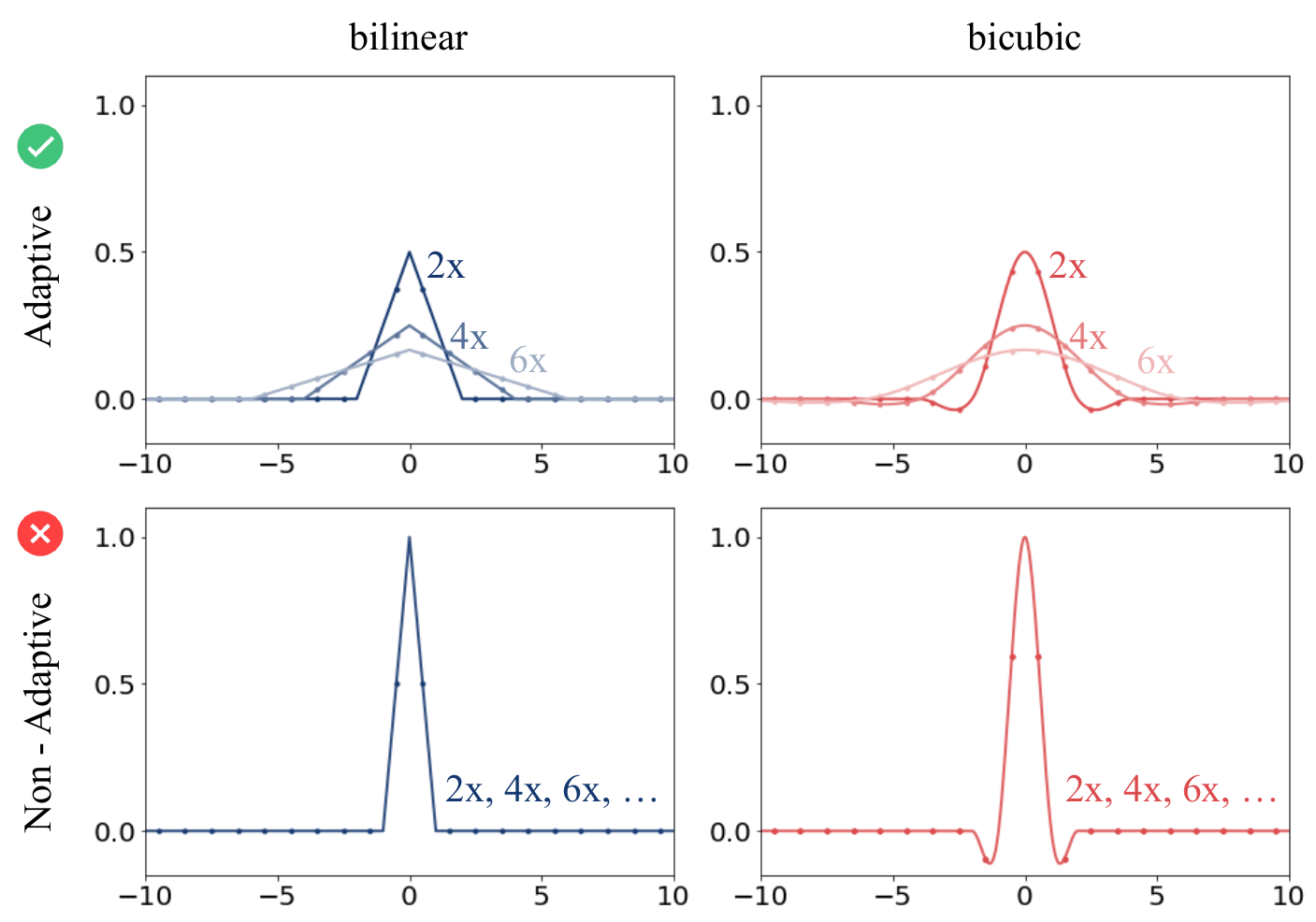}
    \vspace{-7pt}
    \caption{
    \textbf{Interpolation Filters.} \updates{We show the adaptive filters by PIL (top) and non-adaptive filter from PyTorch (bottom). The FID implementations in PyTorch and TensorFlow use a \textit{fixed-width} bilinear interpolation, independent of resizing ratio. In contrast, the proposed Clean-FID uses an implementation that follows standard signal processing principles and adaptively stretches the filter to prevent aliasing. The horizontal axes represent the spatial coordinates and the vertical axes represents the kernel intensity. }
    }
    \lblfig{kernels}
    \vspace{-15pt}
\end{figure}

\section{Introduction}
\lblsec{intro}

With the proliferation of generative modeling techniques, such as Generative Adversarial Networks (GANs)~\cite{goodfellow2014generative}, accurately discerning which methods are performing better has become a critical aspect of the field. For visual data, metrics such as Inception Score (IS)~\cite{salimans2016improved}, Kernel Inception Distance (KID)~\cite{binkowski2018demystifying}, and the ubiquitously-used \fid (FID)~\cite{heusel2017gans} have become standard practice for developing and adopting models. Under the hood, these methods evaluate the discrepancy between generated and natural images, in a deep feature space, to capture relevant features of the two distributions. After all, at its core, generative modeling involves learning and mimicking \textit{high-order, complex} statistics of visual data.

However, we find that \textit{low-level, seemingly innocuous operations}, can induce surprisingly large discrepancies in high-level statistics. For example, consider \reffig{teaser}. Given the same input image, different image processing libraries produce drastically different results. Specifically, the implementations using OpenCV, TensorFlow and PyTorch libraries with default flags, contain severe aliasing artifacts.
Similarly, the simple act of saving images in a JPEG operation with the default parameters, either when building the training dataset or collection of generated images, adds quantization and low-level statistical differences to the underlying data.
The low-level statistical differences induced by these differences cause meaningful variations when used for evaluation protocols.
As the \fid (FID) metric~\cite{heusel2017gans} is the most ubiquitous~\cite{heusel2017gans,karras2019style,brock2018large,razavi2019generating,kingma2018glow}, it is the focus of our experiments. We offer a standard benchmark, \textit{clean-fid} (\url{github.com/GaParmar/clean-fid}), and concrete suggestions on resizing and quantization procedures to enable clean comparisons in future evaluation protocols.

First, we investigate the implications of image resizing. When downsampling, signal processing techniques recommend ``prefiltering'' the input, to prevent high-frequency elements from aliasing into the output. When the downsampling factor is larger, the prefilter kernel should be correspondingly stretched. \updates{However, as shown in \reffig{kernels}, the resizing function used by the FID implementations in TensorFlow and PyTorch \textit{do not} prefilter the image, resulting in aliasing artifacts shown in \reffig{teaser}. }
Resizing can occur in two locations -- during data preprocessing (training with lower resolution) or at evaluation time (resizing to 299 resolution to compute the FID metric). In both cases, inconsistent resizing functions induce variations downstream. If used for data preprocessing, the training data distribution itself is changed. When used for the evaluation metric, small variations in resizing can cause changes in subsequent feature extraction.
We quantify the effects of these inconsistencies and offer standard recommendations. Specifically, we propose to use a stronger bicubic filter \cite{keys81cubic}; more importantly, we propose to adjust prefiltering width based on the resizing factors, as guided by signal processing principles.

Secondly, we investigate the implication of image compression. While the JPEG protocol is a lossy compression scheme, designed to preserve perceptual similarity to the original~\cite{wallace1992jpeg},
it can perturb an image enough to corrupt downstream feature extraction. This affects performance drastically and can create mismatches when comparing methods. Perhaps more surprisingly, when training images are saved with JPEG compression, modern GANs are unable to fully mimic the induced artifacts, and large FID improvements can actually be artificially achieved by tweaking the JPEG compression ratios when storing the generated images. We quantify the surprising effects of this compression operation, and again offer a concrete, standardized protocol to avoid inconsistencies and hindrances to proper evaluation. 

In conclusion, we characterize the surprising importance of
low-level image processing steps, resizing and quantization, when training and evaluating generative models, such as GANs. We focus our experiments on the widely adopted FID metric, and show additional results on the KID metric~\cite{binkowski2018demystifying} as well as IS~\cite{salimans2016improved} and Perceptual Path Length (PPL) metrics~\cite{karras2019style} (in the supplement). Importantly, \textit{any} metric, present or future, that derives statistics from images undergoing these processing steps, will be affected by these factors. More details and results can be found on our \href{https://www.cs.cmu.edu/~clean-fid/}{website}.

\begin{figure*}[ht!]
    \centering
    \includegraphics[width=0.95\linewidth]{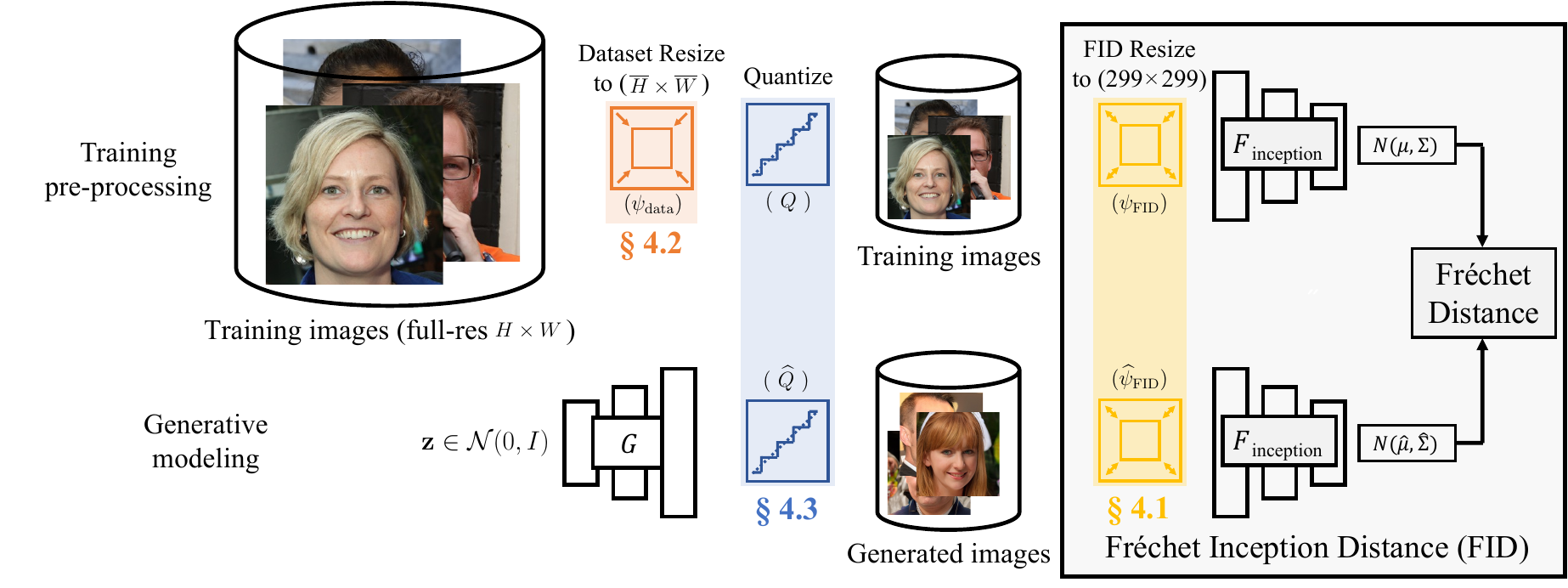}
       \vspace{-10pt}
   \caption{ \textbf{Overview of the steps involved in FID.}
Generative modeling and evaluation require undergo subtleties in image pre-processing. {\bf Top}:  First, the image dataset may be downsampled before training (e.g., 1024$\rightarrow$256 for FFHQ), requiring a resize ($\resizedata$) and possible compression ($\quantizeRef$). {\bf Bottom}:
Generated images may be saved as an unsigned 8-bit integer, resulting in a quantization and possible further compression ($\quantizeTest$).
FID aims to measure how well a generative model $G(z)$ mimics the training distribution. The calculation resizes real and generated images to $299$ resolution ($\resizerealFID$ and $\resizefakeFID$, respectively), extracts deep features using the Inception network~\cite{szegedy2015rethinking}, fits Gaussians, and takes the Fr\'echet distance between two distributions. 
We study the effects of resizing the training images $\resizedata$ in \refsec{resize_fid_jpeg}, resizing to 299$\times$299 $\resizerealFID$ and $\resizefakeFID$ in \refsec{resize_fid} and the quantizations/image compressions $\quantizeTest$ and $\quantizeRef$ in \refsec{resize_data}. 
   }
   \lblfig{fid_steps}
   \vspace{-13pt}
\end{figure*}
\section{Related Work}
\lblsec{related}
\vspace{5pt}
\myparagraph{Deep generative models.}
A wide range of image and video synthesis applications~\cite{zhu2016generative,park2019SPADE,liu2019few,shrivastava2017learning} have been enabled, as a result of tremendous progress in deep generative models such as GANs~\cite{goodfellow2014generative,radford2015unsupervised,karras2018progressive,karras2019style,brock2019large}, VAEs~\cite{kingma2014auto,razavi2019generating, dcvae21}, autoregressive models~\cite{oord2016conditional}, flow-based models~\cite{dinh2017density,kingma2018glow}, and energy-based models~\cite{salakhutdinov2009deep,nijkamp2020anatomy,du2019implicit}. It is often relatively easier to evaluate individual model's performance on downstream computer vision and graphics tasks, as they have a clear target for a given input. However, evaluating unconditional generative models remains an open problem. It is still an important goal, as most generative models are not tailored to any downstream task. 
\vspace{-5pt}
\myparagraph{Evaluating generative models.} The community has introduced many evaluation protocols.  
One idea is to conduct user studies on cloud-sourcing platforms for either assessing the samples' image quality~\cite{denton2015deep,salimans2016improved,zhou2019hype} or identifying duplicate images~\cite{arora2018gans}. 
Due to the subtle differences in user study protocols (e.g., UI design, fees, date/time), it is not easy to replicate results across different papers. Large-scale user studies can also be expensive, prohibiting its usage when evaluating hundreds of model variants and checkpoints during the development stage. Several methods propose evaluating generative models from a self-supervised feature learning perspective, by repurposing the learned discriminators~\cite{radford2015unsupervised} or accompanying encoders~\cite{donahue2019large} for a downstream classification task. However, the representation power of the discriminator or encoder does not directly reflect the generators' sample quality and diversity. In addition, not every generative model is trained with a discriminator or encoder. 

To overcome the previous issues, an area of focus is developing automatic metrics that directly assess the samples of generative models. Various metrics been proposed, criticized, and modified. Commonly-used ones include log-likelihood~\cite{kingma2014auto,goodfellow2014generative}, density estimate with Parzen window~\cite{goodfellow2014generative}, Inception Score~\cite{salimans2016improved}, Perceptual Path Length~\cite{karras2019style}, \fid (FID)~\cite{heusel2017gans},  Classification Accuracy Score and its early variants~\cite{ravuri2019classification,salimans2016improved}, Classifier Two-sample Tests~\cite{lehmann2006testing,lopez2016revisiting}, precision and recall~\cite{sajjadi2018assessing,kynkaanniemi2019improved},  Kernel Inception Distance (KID)~\cite{binkowski2018demystifying}, among others. Each metric has associated pros and cons~\cite{theis2015note,borji2019pros} and none are perfect. 

Among them, \fid (FID) has become the most widely-used metrics, as it can model intra-class diversity better than Inception Score. FID is also easy and fast to compute without training additional classifiers~\cite{ravuri2019classification}, and has been shown to be consistent with human perception~\cite{heusel2017gans}. As a result, it has been used in recent GANs papers~\cite{zhang2019consistency,karras2019style,brock2019large} as well as large-scale evaluation study~\cite{lucic2017gans}, despite facing criticism about the fact that FID is a biased estimator and sensitive to the number of samples used in the evaluation~\cite{chong2019effectively,binkowski2018demystifying}. 
Our goal here is \emph{not} to study which one is a better metric. Instead, we focus our study on the popular FID metric and how subtle details and aliased image resizing functions can affect the final scores. Note that the resizing and quantization we study in are applicable to  any evaluation metric that contains such operations.

\myparagraph{Antialiasing and robustness.} 
The study of resampling signals is central in signal processing~\cite{oppenheim1999discrete}, image processing~\cite{gonzalez2002digital}, and computer graphics~\cite{foley1996computer}. In particular, when downsampling a signal, one must consider the Nyquist sampling criterion~\cite{nyquist1928certain} and antialias to prevent high-frequency information from aliasing into the output. Without proper antialiasing, in the worst case, an adversary can embed a completely different image in the original, resulting in a ``scaling attack''~\cite{xiao2019seeing, quiring2020adversarial}. In convolutional network design, antialiasing has taken form in average pooling~\cite{lecun1998gradient} and Gaussian filtering~\cite{mairal2014convolutional}. While it was replaced by operations such as max-pooling, based on empirical performance~\cite{scherer2010evaluation}, recent works have demonstrated that antialiasing can be compatible and improve performance in convolutional networks~\cite{zhang2019making,zou2020delving}, transformers~\cite{qian2021blending}, NeRFs~\cite{barron2021mip}, and GANs~\cite{karras2021alias}.
Despite these advances, generative methods continue to be detectable~\cite{wang2020cnn, chai2020makes}, and discriminative networks continue to be sensitive to small perturbations, such as shifts~\cite{azulay2019deep,engstrom2019exploring} and JPEG compression~\cite{hendrycks2019benchmarking}. Achieving robustness to such perturbations remains an open problem~\cite{taori2020measuring}, and the preprocessing steps, such as image resizing, used before feature extraction remain consequential. We study the effect of such steps in a generative modeling pipeline and propose a standardization following signal processing principles, in order to facilitate easy and fair comparisons.

\vspace{-1mm}

\section{Preliminaries}
\lblsec{preliminaries}
In this section, we discuss several low-level image processing steps using different popular libraries. 
We find that many of these details can have a large effect on the FID score being computed. \reffig{fid_steps} details the step-by-step process for both dataset preparation and model evaluations.

\subsection{Generative Modeling and Evaluation Pipeline}
\vspace{-5pt}
The \fid (FID) score aims to measure the gap between two data distributions~\cite{heusel2017gans}, such as between a training set and samples from a  generator.
\vspace{-1mm}
\myparagraph{Dataset pre-processing.}
We denote the original real image distribution as $\real \sim \realdist$, where $x \in \mathbb{Z}^{H \times W \times 3}$. Note that images are saved as 8-bit integers, represented by $\mathbb{Z}$. Training and developing large-scale GANs at the original resolution~\cite{brock2019large,karras2019style} is often prohibitively expensive, sometimes requiring training hundreds of models during development. As such, developing on lower-resolution versions of the original dataset is a common practice~\cite{liu2020diverse,zhao2020diffaugment,zhang2020consistency}, such as 1024$\rightarrow$256 on FFHQ or $256\rightarrow128$ on ImageNet.

As shown in the top branch of \reffig{fid_steps}, to prepare a lower-resolution training set, one must downsample the training set, denoted by $\resize_{\text{data}}$. Note that downsampling requires an antialiasing step according to standard textbooks~\cite{oppenheim1999discrete,foley1996computer,szeliski2010computer} that converts integers into a floating point number, $\mathbb{Z} \rightarrow \mathbb{R}$. A quantization step is added afterwards to cast back to $\mathbb{Z}$.
This data preparation step introduces a new data distribution of low-res real images: $\realLR \sim p_{\text{data}}(\realLR)$, where $\realLR \in \mathbb{Z}^{\HLR \times \WLR \times 3}$.
\vspace{-1mm}
\myparagraph{Evaluating a generator with FID.}
A generator $\G$ that learns to map a latent code $\latent \in \mathcal{N}(0, I)$ to output images $\G(\latent) \in \mathbb{R}^{\HLR \times \WLR \times 3}$ is trained on the lower resolution dataset. A common evaluation method is passing both real and generated images through a feature extractor $\F$, fitting a Gaussian distribution, and measuring the Fr\'echet distance between the two distributions. 
Deep network activations are used as the statistics of interest, as they have been shown to correspond well with human perceptual judgments~\cite{zhang2018unreasonable} and are often used as training objectives~\cite{gatys2015neural,johnson2016perceptual,dosovitskiy2016generating}. The feature extractor $\F$ used for this task is an InceptionV3 model~\cite{szegedy2015rethinking}. Because this model is trained on $299\times299\times3$ ImageNet image crops~\cite{deng2009imagenet}, the training and generated images are resized denoted by functions $\resizeFID$ and $\resizefakeFID$, respectively, before being processed. 
As these images may be saved in development pipelines, different image compressions may be applied. These operations are represented by $\quantizeRef$ for reference images $\real$ and by $\quantizeTest$ for synthesized images $\G(\latent)$. 

\vspace{-5pt}
\begin{equation}
    \realfeat = \F(\resizeFID(  \quantizeRef ( \resizedata (\real)))),
\vspace{-2pt}
\lbleq{real_feature}
\end{equation}
\begin{equation}
    \fakefeat = \F ( \resizefakeFID( \quantizeTest (\G(\latent)) ) ).
\lbleq{fake_feature}
\end{equation}
After the images are appropriately resized, and the features are extracted, the mean ($\mu$, $\hat{\mu}$) and covariance matrix ($\Sigma$, $\widehat{\Sigma}$) of the corresponding set of features $\realfeat$ and $\fakefeat$ are used to compute the Fr\'echet distance shown in the equation below. 
\begin{equation}
    \text{FID} = ||\mu - \hat{\mu}||_2^2 + \text{Tr}(\Sigma + \widehat{\Sigma} - 2(\Sigma \widehat{\Sigma})^{1/2}),
\end{equation}
The $\text{Tr}$ operation calculates the trace of the matrix.The different choices for the resizing functions ($\resizedata, \resizeFID, \resizefakeFID$) and quantization functions ($\quantizeRef, \quantizeTest$) adds potential sources of inconsistencies in generative modeling pipelines.

\begin{figure}[t!]
    \centering
    \includegraphics[width=\linewidth]{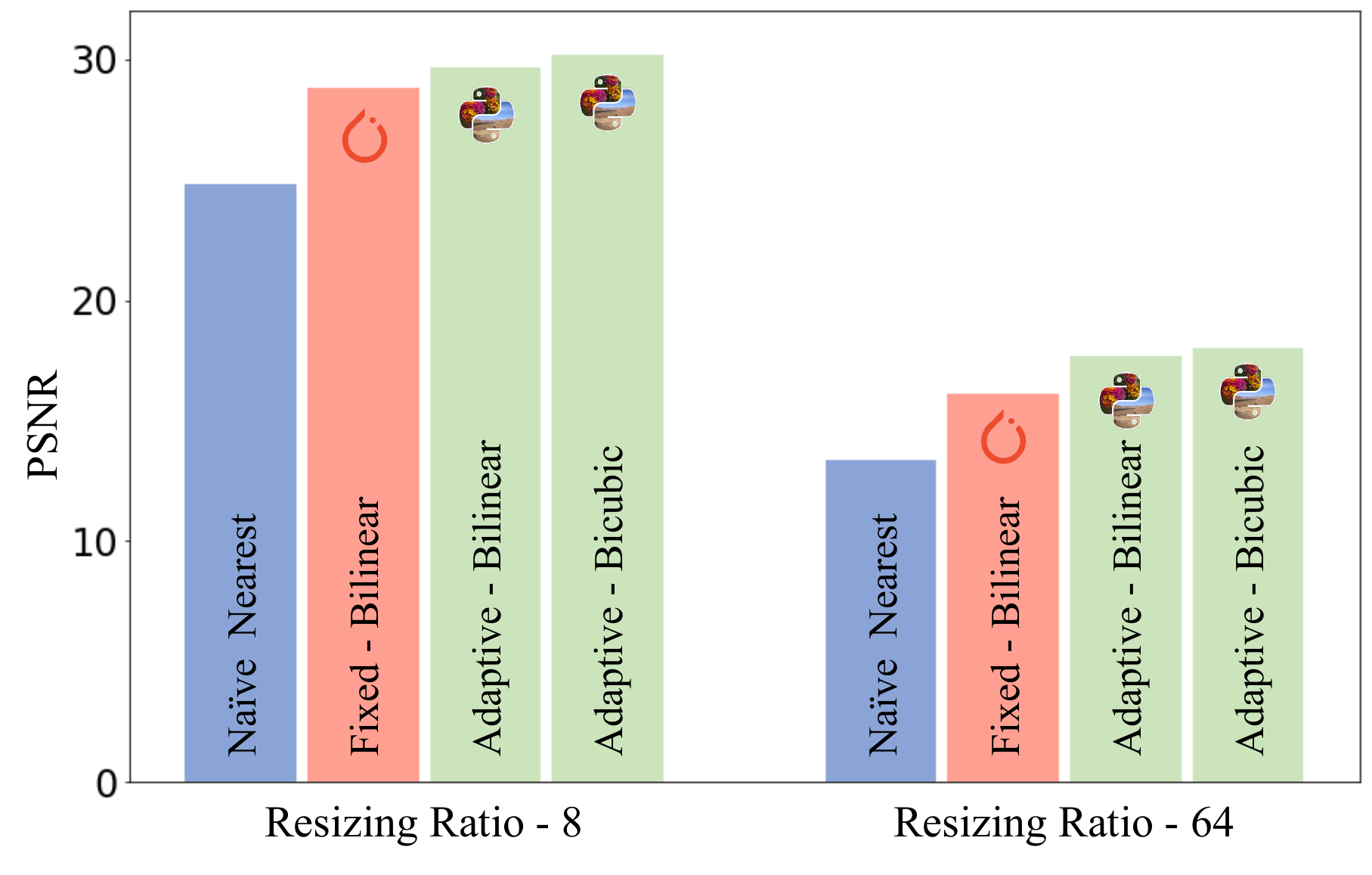}
    \vspace{-9mm}
    \caption{\textbf{Reconstruction after downsampling and upsampling.}
    To illustrate the differences between resizing functions, we downsample images with the different functions and upsample with PIL-Lanczos, and compute similarity to the original with PSNR. The implementation that adjusts prefilter size to downsampling factor (PIL) reconstructs the original more accurately than the implementations that used a fixed filter size (PyTorch).
    This is especially apparent for larger resizing ratios ($64\times$), where performance is closer to naive nearest subsampling.
    }
    \lblfig{up_down}
    \vspace{-5mm}
\end{figure}

\begin{figure*}[t!]
    \centering
    \begin{minipage}{\linewidth}
    \centering
    \includegraphics[width=0.97\linewidth]{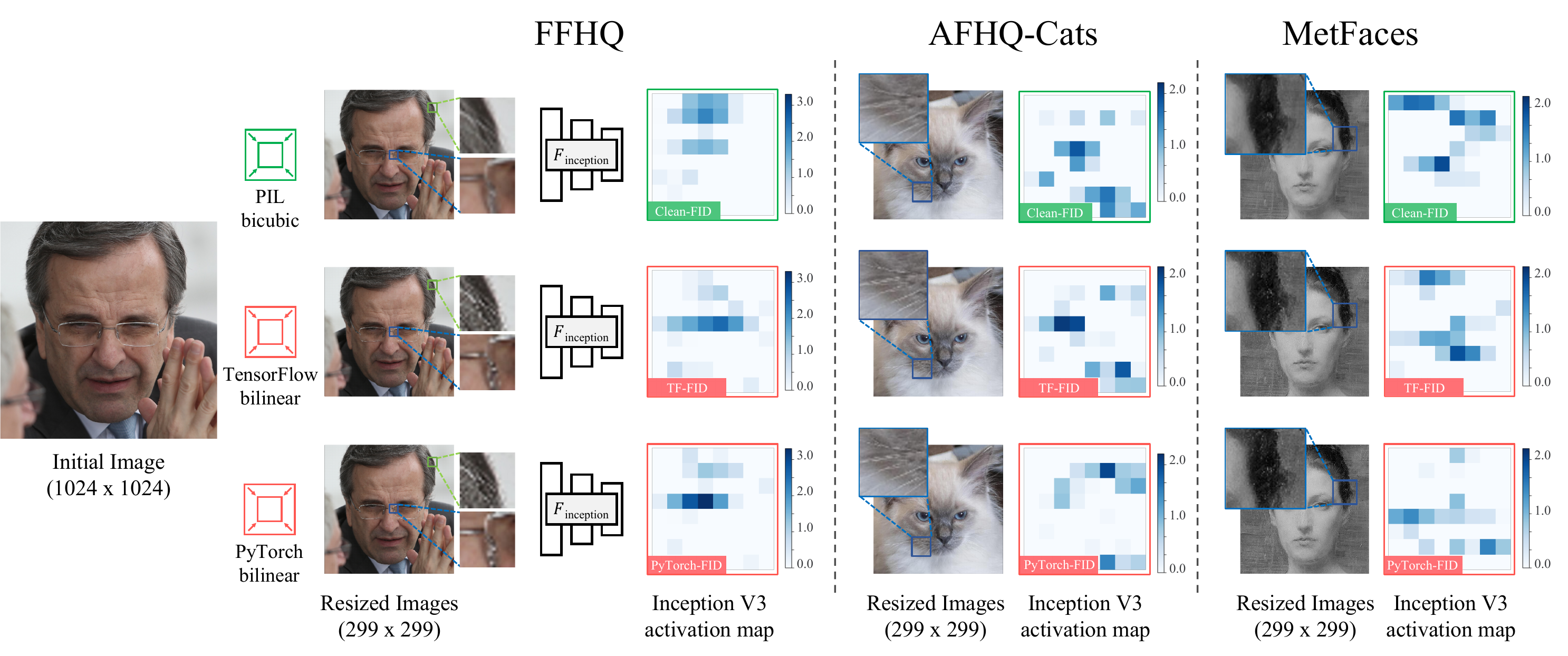}
    \end{minipage}
    \vspace{-3mm}
    \caption{
    \textbf{Differences in Inception features induced by inconsistent resizing.} We resize full resolution $1024 \times 1024$ FFHQ~\cite{karras2019style} image (left) to $299 \times 299$ using PIL-bicubic (top), Tensorflow-bilinear (used by TF-FID) (middle), and Pytorch-bilinear (used by PyTorch-FID) (bottom). The resizing functions using current FID implementations (middle and bottom rows) introduce artifacts; for example, the hair and glasses appear noisier and aliased, as compared to the top row. We observe similar behavior on other commonly-used datasets - AFHQ-Cats ($512 \times 512$) and MetFaces ($1024 \times 1024$). Furthermore, these resizing implementations are inconsistent with each other, inducing different activation maps when passed through the Inception-V3 network~\cite{szegedy2015rethinking}. We propose to resolve this inconsistency and also reduce the aliasing, by standardizing bicubic downsampling as the preprocessing function for a ``Clean-FID'' (using filtering that adjusts to the downsampling factor, adhering to signal processing principles).
    }
   \lblfig{feature_maps}
   \vspace{-5pt}
\end{figure*}

\subsection{Image Resizing}
\lblsec{image_resizing}
\vspace{-3pt}
Depending on the dataset and training size, the resizing operations ($\resizeFID$, $\resizefakeFID$) in \reffig{fid_steps} can either be downsampling
or upsampling.
Downsampling is the primary focus of this investigation, as it involves \textit{throwing away information}. Methods for downsampling is a common study in the fields of signal and image processing~\cite{oppenheim1999discrete,gonzalez2002digital}.

\myparagraph{Antialiasing by prefiltering.}
The most naive approach is to simply subsample (taking every N$^\text{th}$ element if performing downsampling by an integer factor N), sometimes referred to as \textit{nearest}. This corresponds to filtering the input image with Kronecker delta function, as only a single value is drawn. Such an approach leads to aliasing, as high-frequency elements of the input alias to the output.

A central principle in image processing, signal processing, graphics, and vision~\cite{foley1994introduction,forsyth2012computer,szeliski2010computer,gonzalez2002digital, oppenheim1999discrete} is to blur or ``prefilter'' before subsampling, as a means of removing high-frequency information (thus preventing its misrepresentation downstream). For linear filters, this corresponds to a ``depth-wise convolution'', using deep learning parlance~\cite{sifre2014rigid,howard2017mobilenets}. We explain two important ways in which prefiltering implementations can vary.

\myparagraph{Filter size adaptation to downsampling factor.} First, according to signal processing principles, the size of the filter \textit{should} be adjusted, in accordance with the downsampling factor. Widening the low-pass filter in the spatial domain corresponds to reducing its bandwidth and filtering more aggressively in frequency space. As a larger downsampling factor means a lower bandwidth can be represented on the output signal, \textit{widening the filter accordingly is necessary to prevent aliasing}. However, in many common implementations, this is not implemented (or is not used by default); instead, a filter of \textit{fixed, non-adaptive} size is used.
\vspace{-1mm}
\myparagraph{Choice of filters.} Secondly, there is a choice of different convolutional filters. The idealized low-pass filter is a sinc, requiring infinite support. As such, approximate filters with different subtle tradeoffs in runtime and behavior are used instead. The \textit{box}, also known as \textit{area} filter, corresponds to a rectangular filter, computing the average of values within a neighborhood. The \textit{bilinear} filter is a triangular filter, \textit{bicubic}~\cite{keys81cubic} is a stronger cubic function, and the \textit{lanczos} filter is an enveloped sinc. All perform a weighted average and have stronger antialiasing, closer to the idealized sinc. \updates{See Appendix~\ref{sec:ap_filters} for additional details about the different interpolation filters. }

\vspace{-1mm}
\myparagraph{Practical implications of implementation variations.} We investigate the inconsistencies that can arise, when these two factors are varied, and show a toy example in \reffig{teaser} in downsampling a circle. While the choice of filter is largely constant across libraries (lanczos, bicubic, bilinear are shown in each column), \textit{the choice of whether the filter adapts to the downsampling factor is not}. While the PIL library adapts the filter (top row), other libraries do not by default, leading to aliased results. In particular, FID implementations of TensorFlow-FID and PyTorch-FID, use bilinear downsampling implementations that exhibit aliasing, and thus are the focus of our study.

An implication of aliasing is a suboptimal representation of the original image. In Figure~\ref{fig:up_down}, we show the result of downsampling and upsampling an image, and comparing it to the original with PSNR (averaged over 300 FFHQ images). The methods with non-adaptive filters achieve a worse reconstruction than a method that adapts the filter. This effect is more significantly accentuated with larger downsampling factors, where high-frequency aliasing dominates when using non-adaptive filters. \reffig{feature_maps} shows how the Inception features are affected by aliased resizing functions for various datasets. 

\myparagraph{Recommendation.} Above, we have established that the implementations of FID are inconsistent and aliased. Ideally, the community can (a) use a consistent pipeline to facilitate fair comparisons across papers, and (b) follows signal processing principles and antialiases, in order to best represent the underlying data it is trying to characterize. We propose to use an adaptive filter (and thus produce consistently antialiased results). Second, we propose to use a bicubic, instead of bilinear filter, which offers stronger reconstruction. While such an implementation is currently found in PIL, future implementations that are computationally equivalent would be of use).

\subsection{Quantization and Image Compression}
\lblsec{image_format}

\begin{figure}[tb!]
    \centering
    \includegraphics[width=.9\linewidth]{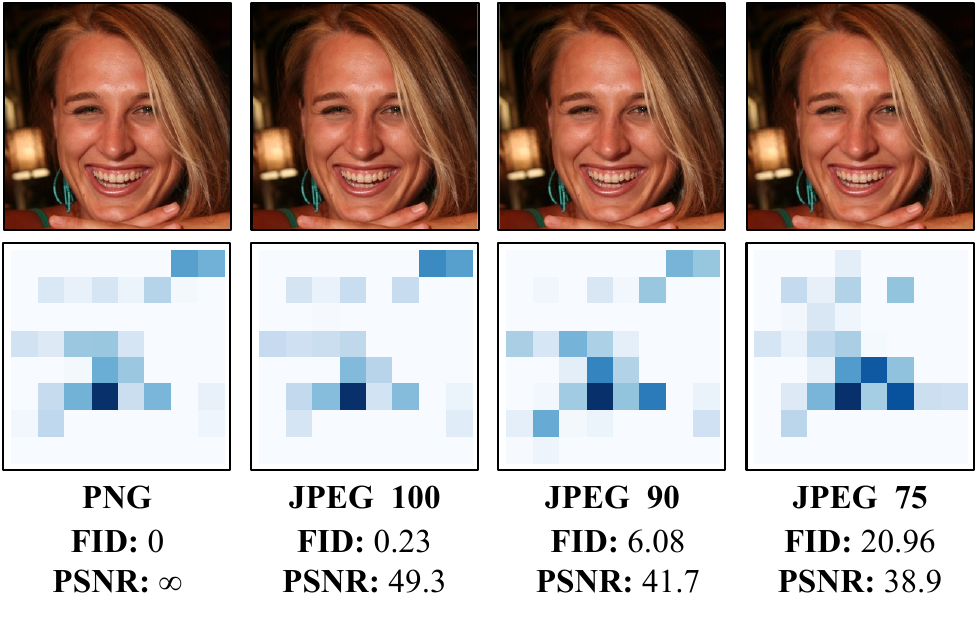}
    \vspace{-2mm}
    \caption{\textbf{Effects of JPEG compression on an image.} We show a sample image from the FFHQ dataset \cite{karras2019style}, saved with lossless PNG and different JPEG compression ratios. The FID scores under the images are calculated between FFHQ images saved using the corresponding JPEG format and the PNG format. PSNR is computed with 1000 images. While the images are perceptually similar, this induces changes in the Inception-V3 activations, resulting in large FID scores. }
    \lblfig{jpeg_faces}
           \vspace{-5pt}
\end{figure}
\vspace{5pt}
\myparagraph{8-bit Quantization.}
While images are represented by 8-bit integers $\mathbb{Z}$, operations such as resizing and data augmentation, as well as the raw generator output will provide floating point numbers $\mathbb{R}$. Post-processing the results introduces more subtleties and affects standard metrics such as FID.
Most simply, an image can be quantized by clipping the output between $[0,255]$ and rounding to produce integers. This is a lossy step and only done when images need to be saved. 
Additionally, we observe that performing this step has a minor effect on the FID score ($<0.01$). 
\vspace{-2mm}
\myparagraph{Image compression.} Saving the image as a raw matrix of values is data-intensive. However, an image contains redundant information that can be exploited. For example, the PNG format compresses an image losslessly.
To further save storage, images are commonly saved using the JPEG codec. While JPEG is a lossy compression technique, it aims to make changes that the human visual system is less sensitive to, namely reducing information in higher frequencies and chroma (color) components~\cite{wallace1992jpeg}. JPEG converts an image into a YCbCr space, subsamples the chroma components, divides images into 8$\times$8 blocks, computes the Discrete Cosine Transform (DCT), and performs quantization. The quantization step facilitates a trade-off between the fidelity of the original image and the amount of the storage saved. In the PIL implementation~\cite{clark2015pillow}, this is done using a ``quality'' option (0-100), which linearly scales the quantization tables (which controls which frequencies are quantized to what granularity).
\updates{Note that setting the quality flag to 100 is \textit{not} a lossless operation. Even when the quantization tables are not scaled, the DCT coefficients are quantized to integer values and the chroma components are subsampled. }

\myparagraph{Image compression changes deep network activations.} In \reffig{jpeg_faces}, we show a real image sampled from the FFHQ dataset~\cite{karras2019style} at a resolution of 256, saved with lossless PNG and lossy JPEG (quality flags set to 100, 90, and 75). Despite being perceptually indistinguishable (with high PSNR values of $\ge 39$), the FID scores increase. The PIL default of 75 results in a high score (21), for example. Note that this FID score is far higher than the score from a powerful generative model, StyleGAN2~\cite{karras2020analyzing} (around 3). Also, variations across recent methods are typically within $1$ FID on FFHQ. We further investigate the implications of using JPEG compression in various parts of the pipeline in experiments below.

\section{Experiments}
\lblsec{expr}
In \refsec{preliminaries}, we outlined the various image processing steps involved in generative modeling pipelines and evaluation. In this section, we introduce sources of variation at these steps and empirically quantify their impacts. 
As depicted in \reffig{fid_steps}, the variations in the FID score arises from three distinct steps: resizing in the FID evaluation step ($\resizerealFID$, $\resizefakeFID$), resizing in the data preprocessing step ($\resizedata$), and quantizing of images ($\quantizeRef$, $\quantizeTest$). We investigate each of these steps in \refsec{resize_fid}, \refsec{resize_data}, and \refsec{resize_fid_jpeg} respectively. 

\subsection{Variation due to FID Resizing}
\lblsec{resize_fid}

Here we investigate the effects of different resizing methods ($\mathbf{\resizerealFID}$, $\mathbf{\resizefakeFID}$) used in the  FID calculation step.  

\myparagraph{Variation induced by resizing functions on real images.} 
We start with two sets of full-resolution $1024 \times 1024$ face images - from the FFHQ dataset, and from a pre-trained StyleGAN2 generator. Each of the sets of images is resized from 1024$\rightarrow$299 using different methods. In \reftbl{fid_resize} (left), we compare the set of real images resized with the antialiased resizing operation (PIL bicubic) to the \textit{same set of real images}, resized using other aliased functions that use a fixed width prefiltering kernel. As we compare the same set of images, we anticipate all FID and KID scores to be close to 0 and the PSNR values to be very high. However, as shown in Figures~\ref{fig:teaser}, \ref{fig:kernels}, and \ref{fig:feature_maps}, only a subset of the commonly used resizing operators adjust the filter width and antialias the images. These differences in resizing operations cause drastic changes in the Inception-V3~\cite{szegedy2015rethinking} activation maps. 

Filters that adapt their size and antialias are more consistent, even with different filter types -- PIL-bilinear has FID 0.64 as compared to PIL-bicubic.
On the other hand, implementations that ignore the downsampling factor (PyTorch and TensorFlow) show much larger deviation (FID 4.3), with scores nearing naive nearest (FID 7.4), that does not filter at all. This indicates that whether the filter adapts to the downsampling filter can change the modeled data distribution by non-trivial amounts.

\myparagraph{Variation induced by resizing functions on generated images.}
After studying the effects on real images, we evaluate how different resizing function $\resizefakeFID$ choices affect the FID score when used in a full generative modeling pipeline.
Here, we evaluate a pretrained StyleGAN2 generator~\cite{karras2020analyzing} trained on FFHQ (1024), MetFaces (1024), and AFHQ (512) dataset images, and calculate FID with 50,000 images.
In \reftbl{fid_resize} (right), we consider the asymmetric case, where features for the real images and generated images use different resizing functions. This case arises when features for real images are pre-computed and shared by one group of authors, while generated features may be calculated on the fly with a different library.
Here, we observe that using the same resizing function as the reference dataset (PIL-bicubic) achieves the lowest performance. Using a different resize function, such as PIL-bilinear increases the score to 4. Using an aliased function increases the score drastically to 7, close to naive subsampling ($>10$).

Next, in \reftbl{fid_resize_same}, we show a comparison when the same resizing function is used for the real dataset images and the StyleGAN2 generated images. Interestingly, we observe that the \textit{aliased} resizing functions result in lower FID scores across multiple commonly used datasets - FFHQ (1024), MetFaces~\cite{karras2020training} (1024), and AFHQ~\cite{choi2020starganv2} (512).
This indicates that using the antialiased function as preprocessing makes the downstream FID calculation more sensitive at measuring the discrepancies between distributions.
\subsection{Variation due to Dataset Resizing}
\lblsec{resize_data}

Previously, we considered the scenario when the dataset was not downsampled.
However, as discussed in \refsec{intro} and illustrated in \reffig{fid_steps}, dataset downsampling is needed when training a model on a low-resolution version of the original dataset~\cite{zhao2020diffaugment,karras2020training,zhang2020consistency}
(e.g., $256$ for FFHQ or $128$ for ImageNet).
Before, the target distribution was fixed, and differences were purely introduced during post-hoc metric evaluation. Now, the situation is much more intricate. \textit{Different resizing choices will result in different training distributions entirely}.

In \reftbl{fid_256_models}, we train three different StyleGAN2 \cite{karras2020analyzing} (config-e) models, following the official PyTorch implementation\footnote{https://github.com/NVlabs/stylegan2-ada} for 25k iterations. We resize FFHQ \cite{karras2019style} to $256$ using Naive Nearest, PIL--bicubic, PyTorch--bilinear, and TensorFlow--bilinear.
We use the same PIL--bicubic function ($\resizerealFID$, $\resizefakeFID$) for FID evaluation; note that here, it is upsampling ($256\rightarrow299$).
Qualitatively, using an aliased downsampling function produces a training distribution with visual artifacts for the generative model to mimic, likely different than the natural visual data we wish to model.
Quantitatively, interestingly, we observe that that the aliased pre-processing results in \textit{lower} FID values. As the antialiased function better preserves signal in the original images, we hypothesize that retaining more information from the original input actually produces a more difficult distribution to model.

\begin{table}[t!]
    \centering
    \resizebox{\linewidth}{!}{
    \begin{tabular}{l@{\hskip 3pt} ccc cc}
        \toprule 
        & \multicolumn{5}{c}{\textbf{PIL--bicubic(Real Images) vs.}} \\
        \cmidrule(lr){2-6}
        \multirow{3}{*}{\textbf{Resize function}} &
        \multicolumn{3}{c}{\textbf{Resize(Real Images)}} &
        \multicolumn{2}{c}{\textbf{Resize(StyleGAN2)}} \\ 
        \cmidrule(lr){2-4} \cmidrule(lr){5-6}
        
        & \multirow{2}{*}{\shortstack[c]{\textbf{FID} \\ $\downarrow$}} & \textbf{KID} & \textbf{PSNR} & \multirow{2}{*}{\shortstack[c]{\textbf{FID} \\ $\downarrow$}} & \multirow{2}{*}{\shortstack[c]{\textbf{KID} \\ $\times10^3\downarrow$}} \\
        & & \textbf{$\times10^3\downarrow$} & [db] $\uparrow$ & \\
        \hline %
        PIL--bicubic         (\IconCheck) & 0   & 0 & $\infty$ & 2.98 & 0.51 \\
        PIL--bilinear        (\IconCheck)  & 0.64 & 0.61 &  45.7 &  4.03 & 1.52 \\
        TensorFlow--bilinear (\IconCross) & 4.34 & 4.32 & 37.66 &  7.45 & 5.12 \\
        PyTorch--bilinear (\IconCross) & 4.36 & 4.31 & 37.66 &  7.45 & 5.15 \\
        Naive nearest         (\IconCross) & 7.43 & 7.54 & 35.16 & 10.67 & 8.47 \\
        \bottomrule 
    \end{tabular}}
    \vspace{-5pt}
    \caption{
    \textbf{Deviations induced by varying resizing implementations.} We measure the discrepancy between real images downsampled with PIL-bicubic ($1024\rightarrow299$) vs. other downsampling functions ($\mathbf{\resizefakeFID}$) on the left. If all downsampling functions were equivalent, the neural metrics (FID \& KID) should be 0 and PSNR $\infty$.
    PIL--bilinear and bicubic adjust antialiasing to the downsampling factor (\IconCheck) and produce relatively low neural metric scores and high PSNRs. Functions using fixed width filters (\IconCross) produce higher discrepancies. Naive nearest does not antialias at all.
    A similar trend holds on synthetic StyleGAN2~\cite{karras2020analyzing} images.
    }
    \lbltbl{fid_resize}

\end{table}

\begin{table}[t!]
    \centering
    \resizebox{\linewidth}{!}{
    \begin{tabular}{l cccc}
        \toprule 
        \multirow{3}{*}{\textbf{Resize function}} & 
        \multicolumn{4}{c}{\textbf{Resize(Dataset Images) vs. Resize(StyleGAN2)}} \\
        \cmidrule(lr){2-5}
                
         & \textbf{FFHQ} & \textbf{MetFaces} & \textbf{AFHQ-Cats} & \textbf{AFHQ-Dogs} \\ 
        
        & \textbf{FID} $\downarrow$ & \textbf{FID} $\downarrow$ & \textbf{FID} $\downarrow$ & \textbf{FID} $\downarrow$ \\
        \hline 
        PIL--bicubic         (\IconCheck) & 2.98 & 65.32 & 5.13 & 20.16 \\
        PIL--bilinear        (\IconCheck) & 2.99 & 64.31 & 5.01 & 19.60 \\
        TensorFlow--bilinear (\IconCross) & 2.75 & 57.45 & 4.93 & 19.45 \\
        PyTorch--bilinear (\IconCross) & 2.75 & 57.46 & 4.94 & 19.46 \\
        Naive nearest         (\IconCross) & 2.68 & 55.09 & 4.80 & 18.25 \\
        \bottomrule 
    \end{tabular}}
    \vspace{-5pt}
    \caption{
    \textbf{Resizing functions affect FID scores.} Here, both resizing functions on real and synthetic images ($\mathbf{\resizerealFID}$, $\mathbf{ \resizefakeFID}$) are the same as each other. If all resizing functions were consistent, all rows would be equal. 
    Interestingly, the downsampling methods that alias result in lower scores; the lowest score is achieved by naive nearest subsampling. Methods that adjust the prefilter size to downsampling factor (implemented by PIL) better preserve information of the original images. This indicates that antialiasing enables subsequent FID to more sensitive to differences in the distributions.}
    \lbltbl{fid_resize_same}
    \vspace{-10pt}
\end{table}

\begin{table}[t!]
    \centering
    \scalebox{0.9}{
    \begin{tabular}{l c}
        \toprule
        \multirow{2}{*}{\textbf{Dataset preprocessing}} & \textbf{FID $\downarrow$ on FFHQ} \\
        & PIL-bicubic \\
        \hline 
        Naive Nearest (\IconCross)        & 4.82 $\pm$ 0.09\\
        PyTorch--bilinear (\IconCross)    & 5.13 $\pm$ 0.20\\
        TensorFlow--bilinear (\IconCross) & 5.08 $\pm$ 0.16\\
        PIL--bicubic (\IconCheck)         & 6.21 $\pm$ 0.23\\
        
        \bottomrule
    \end{tabular}}
        \vspace{-5pt}
    \caption{ \textbf{Dataset resizing.}  We downsample the FFHQ dataset using different resize functions $\mathbf{\resize_{data}}$ from 1024 to 256. We train StyleGAN2 \cite{karras2020analyzing} (Config-E) models, using the identical training procedure and report FID of the result. \updates{The score is computed across three different training runs for each of the setting.} The scores show large variation, indicating the resizing function can greatly affect the training distribution. Using a preprocessing function that antialiases (marked by \IconCheck) preserves more information from the original images and interestingly results in a higher score. 
    }

    \lbltbl{fid_256_models}
    \vspace{-10pt}
\end{table}

\begin{figure}[tb!]
    \centering
    \includegraphics[width=.95\linewidth]{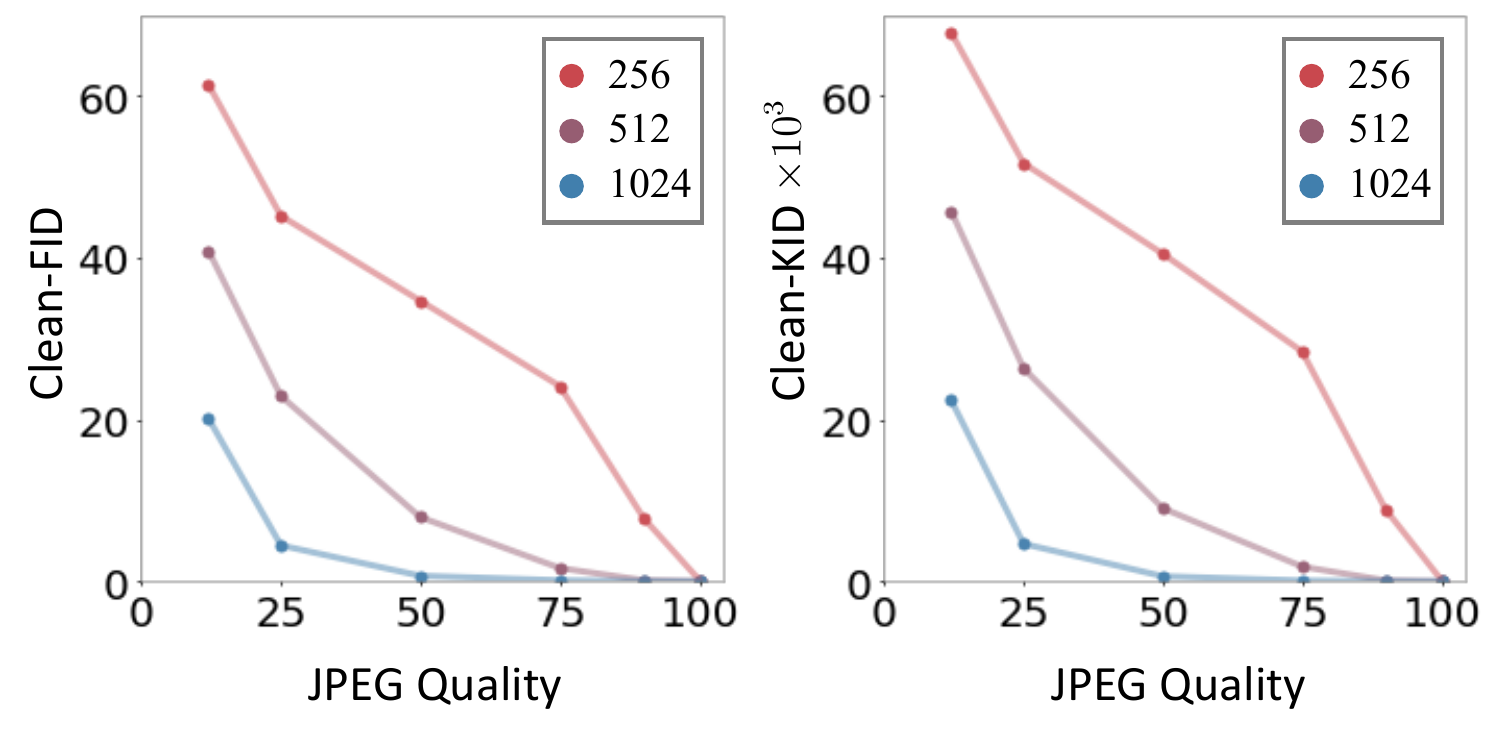}
    \vspace{-3mm}
    \caption{\textbf{Effects of JPEG compression on evaluation metrics.} \updates{
    The FFHQ dataset images are resized from 1024 to different resolutions (512 and 256) using PIL-bicubic and compressed using the JPEG format, with different compression ratios. Subsequently,
    we plot the FID (left) and KID (right) between the compressed images and uncompressed images, at the same resolution, as a function of JPEG compression. The effect of JPEG compression is increasingly more severe for smaller images. } 
    }
    \lblfig{jpeg_effects}
           \vspace{-5pt}
\end{figure}

\begin{figure}[t!]
    \centering
    \includegraphics[width=\linewidth]{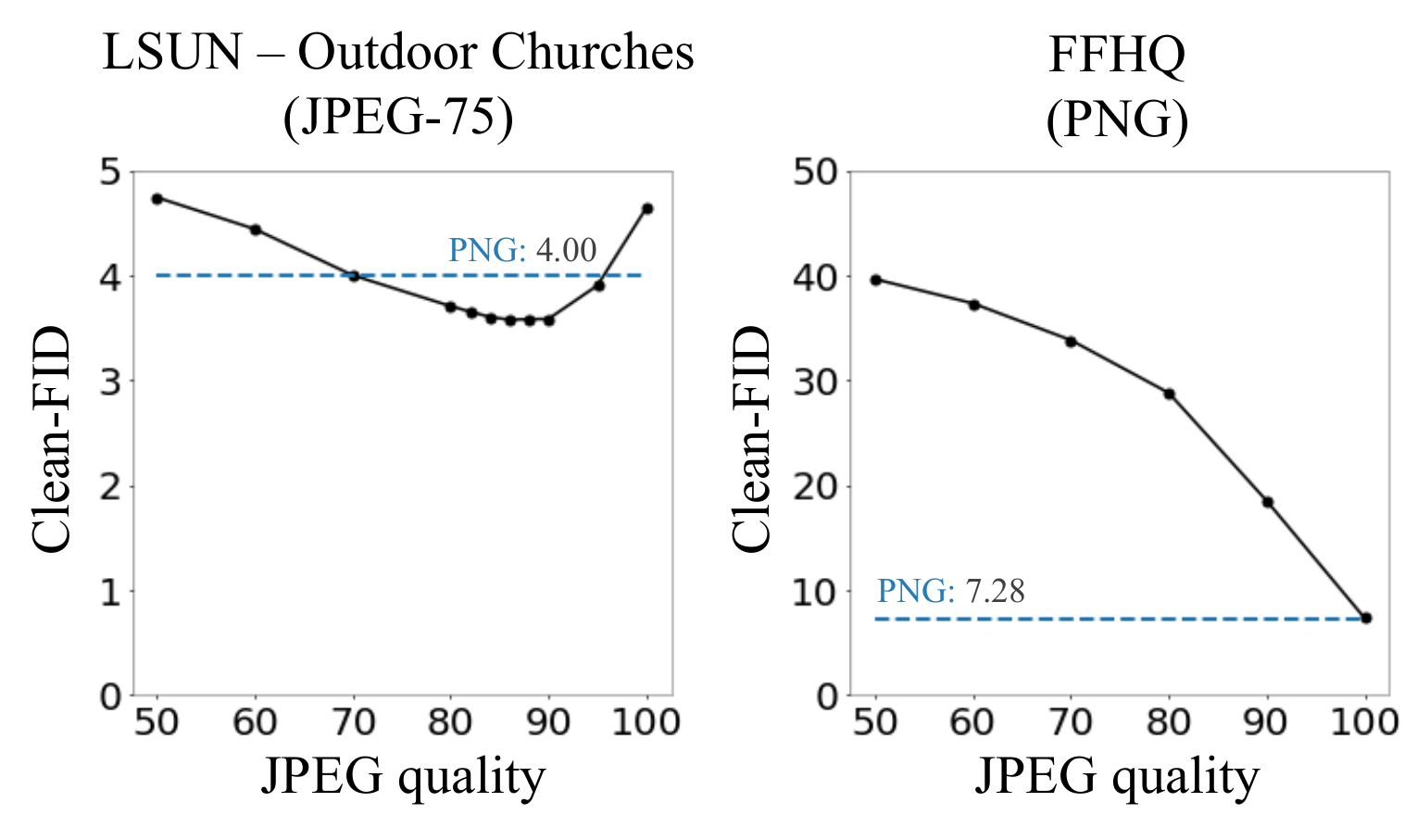}
    \vspace{-20pt}
   \caption{\textbf{Effects of image quantization/compression.} 
   We plot FID as a function of JPEG compression, applied to StyleGAN2 images~\cite{karras2020analyzing}, trained on LSUN Churches~\cite{yu15lsun} (left) and FFHQ~\cite{karras2019style} (right) at a resolution of $256 \times 256$.
   The blue dashed line shows FID when the generated images are quantized to 8-bit unsigned integers (PNG).
   Interestingly, when training with JPEG-75 dataset images (left), applying lossy compression artifically improves the FID score by a large margin (4.00$\rightarrow$3.48).
}
   \lblfig{jpeg_effect}
  \vspace{-5pt}
\end{figure}

\begin{figure}[t!]
    \centering
    \includegraphics[width=\linewidth]{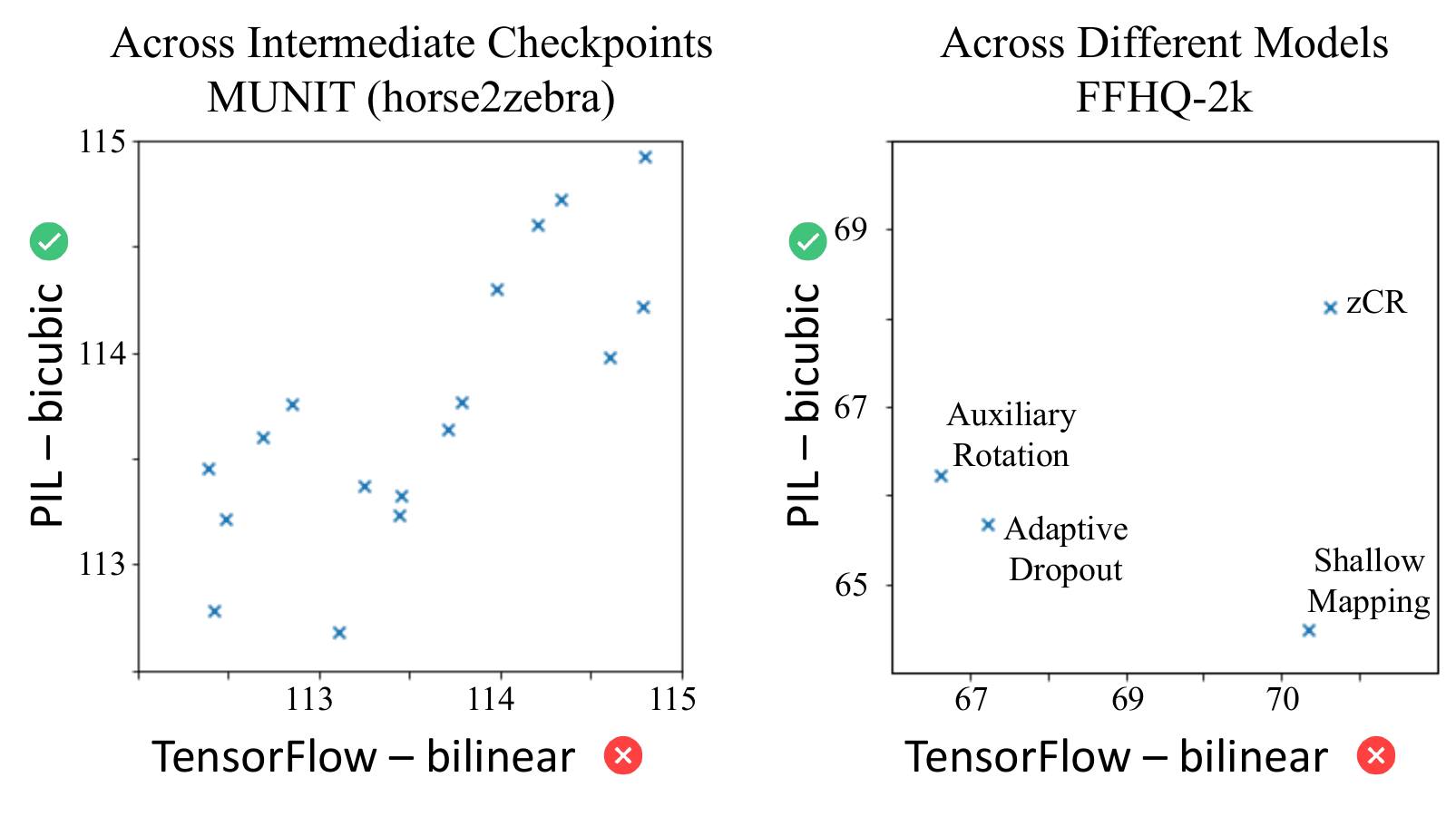}
  \vspace{-8mm}
  \caption{
  \textbf{FID inconsistencies when evaluating models and checkpoints.} We compare the FID scores induced by different resizing functions. (Left) We show different intermediate checkpoints while training a MUNIT model \cite{huang2018multimodal} on the horse2zebra dataset~\cite{zhu2017unpaired}. (Right) We compare methods trained on FFHQ-2k. The non-monotonic relationship demonstrates the sensitivity of the FID metric to the resizing function. As a consequence, different checkpoints or methods may be selected, depending on if an aliased or an anti-aliased resizing function is chosen.}
  \lblfig{selection}
  \vspace{-5mm}
\end{figure}

\subsection{Variation due to Quantization/Compression}
\lblsec{resize_fid_jpeg}

\myparagraph{JPEG during evaluation.} 
\updates{In \reffig{jpeg_effects}, we test the effect of quantization applied to real FFHQ images at different resolutions on the FID (left) and KID (right) metrics. For each resolution, the real dataset images are correspondingly downsampled using PIL--bicubic, and the scores are computed between the resized uncompressed PNG images and the resized JPEG-compressed images.
\reffig{jpeg_effects} shows that the effect of the JPEG compression on both metrics. The effect is more pronounced for lower resolutions, where the artifacts remain after the subsequent resampling step. 
}

\vspace{-2mm}
\myparagraph{JPEG on training images.} In both comparisons above, each method was compared with the FFHQ dataset images, which were collected as uncompressed PNG files. Any additional compression only monotonically increases the FID score (\reffig{jpeg_effect} right). This is expected, as information is being removed from the generator.

However, this does not apply to other datasets which were collected as JPEG images. 
To study this effect, we train a StyleGAN2 model~\cite{karras2020analyzing} on the LSUN outdoor Church dataset~\cite{yu15lsun}, which saved as JPEG-75 images during data collection. In \reffig{jpeg_effect} (left), we plot the FID of the trained generator as a function of JPEG compression. Surprisingly, we observe that the FID score for the StyleGAN2 model
\textit{actually improves when slight JPEG compression is added}. This indicates that interestingly, though the model is able to capture complex variations in the dataset, it is unable to fully model the low-level statistics induced by JPEG compression. Interestingly, the best FID score (3.48) is obtained when the generated images are compressed with JPEG quality 87 (not the full 75), indicating the model is able to replicate some of the artifacts, but not all. The FID score for the generated images stores as PNG files is 4.00. Furthermore, this indicates that the metric is sensitive to low-level statistics, and a large gain in the metric could be achieved simply through manual post-processing. Following these observations, we recommend that researchers curate and store training images as PNG formats for the future image synthesis datasets.  

\subsection{Consequences in model selection}
\lblsec{consequences}
In this section, we show that using an aliased, as opposed to antialiased implementation can result in different conclusions, both when comparing across different methods and when choosing a ``best'' model checkpoint.
In particular, in \reffig{selection} (left) we evaluate the different intermediate checkpoints when training an image-to-image translation model \cite{huang2018multimodal} on the horse2zebra dataset. In \reffig{selection} (right) we evaluate the StyleGAN2~\cite{karras2020analyzing} models with different data augmentation trained to generate $256 \times 256$ FFHQ images \cite{karras2019style} in a few shot setting (2000 training images). Note that using an aliased resizing implementation for computing the FID metric and choosing the best model can lead to a different best model getting selected.

\section{Recommendations}
\lblsec{recommendation}
We have shown surprisingly large sensitivities to seemingly inconsequential implementation details when evaluating generative models. The resize operation and the image quantization/compression are especially impactful. Based on our observations, we discuss some best practices when training and evaluating a generative model.
We recommend using implementations that adapt the filter size to the downsampling factor, following signal processing principles, at each of the resizing steps ($\resize_\text{data}$, $\resizeFID$, and $\resizefakeFID$) involved. 
There are many details one needs to keep track of when computing the FID score. Any inconsistency in the steps leads to results that are no longer comparable to other methods. To facilitate an easy comparison, avoid inconsistent comparisons, and encourage the usage of critical operations that are correctly implemented, we provide an easy-to-use library, \reponame, at \url{github.com/GaParmar/clean-fid} and pre-computed statistics of Inception features for commonly used datasets.

\vspace{10pt}
\small{
\noindent \textbf{Acknowledgments.} 
We thank Jaakko Lehtinen and Assaf Shocher for bringing attention to this issue and for helpful discussion. We thank Sheng-Yu Wang, Nupur Kumari, Kangle Deng, and Andrew Liu for useful discussions. We thank William S. Peebles, Shengyu Zhao, and Taesung Park for proofreading our manuscript. We are grateful for the support of Adobe, Naver Corporation, and Sony Corporation. 
}

\newpage
\typeout{}
{\small
\bibliographystyle{ieee}
\bibliography{main}

\begin{thebibliography}{10}\itemsep=-1pt

\bibitem{abadi2016tensorflow}
Mart{\'\i}n Abadi, Paul Barham, Jianmin Chen, Zhifeng Chen, Andy Davis, Jeffrey
  Dean, Matthieu Devin, Sanjay Ghemawat, Geoffrey Irving, Michael Isard, et~al.
\newblock Tensorflow: A system for large-scale machine learning.
\newblock In {\em 12th $\{$USENIX$\}$ symposium on operating systems design and
  implementation ($\{$OSDI$\}$ 16)}, pages 265--283, 2016.

\bibitem{arora2018gans}
Sanjeev Arora and Yi Zhang.
\newblock Do gans actually learn the distribution? an empirical study.
\newblock In {\em International Conference on Learning Representations (ICLR)},
  2018.

\bibitem{azulay2019deep}
Aharon Azulay and Yair Weiss.
\newblock Why do deep convolutional networks generalize so poorly to small
  image transformations?
\newblock {\em Journal of Machine Learning Research}, 20:1--25, 2019.

\bibitem{barron2021mip}
Jonathan~T. Barron, Ben Mildenhall, Matthew Tancik, Peter Hedman, Ricardo
  Martin-Brualla, and Pratul~P. Srinivasan.
\newblock Mip-nerf: A multiscale representation for anti-aliasing neural
  radiance fields.
\newblock {\em ICCV}, 2021.

\bibitem{binkowski2018demystifying}
Miko{\l}aj Bi{\'n}kowski, Danica~J Sutherland, Michael Arbel, and Arthur
  Gretton.
\newblock Demystifying mmd gans.
\newblock In {\em ICLR}, 2018.

\bibitem{borji2019pros}
Ali Borji.
\newblock Pros and cons of gan evaluation measures.
\newblock {\em Computer Vision and Image Understanding}, 179:41--65, 2019.

\bibitem{bradski2000opencv}
Gary Bradski and Adrian Kaehler.
\newblock Opencv.
\newblock {\em Dr. Dobb’s journal of software tools}, 3, 2000.

\bibitem{brock2018large}
Andrew Brock, Jeff Donahue, and Karen Simonyan.
\newblock Large scale gan training for high fidelity natural image synthesis.
\newblock In {\em International Conference on Learning Representations (ICLR)},
  2019.

\bibitem{brock2019large}
Andrew Brock, Jeff Donahue, and Karen Simonyan.
\newblock Large scale gan training for high fidelity natural image synthesis.
\newblock In {\em International Conference on Learning Representations (ICLR)},
  2019.

\bibitem{chai2020makes}
Lucy Chai, David Bau, Ser-Nam Lim, and Phillip Isola.
\newblock What makes fake images detectable? understanding properties that
  generalize.
\newblock In {\em European Conference on Computer Vision}, pages 103--120.
  Springer, 2020.

\bibitem{chen2015mxnet}
Tianqi Chen, Mu Li, Yutian Li, Min Lin, Naiyan Wang, Minjie Wang, Tianjun Xiao,
  Bing Xu, Chiyuan Zhang, and Zheng Zhang.
\newblock Mxnet: A flexible and efficient machine learning library for
  heterogeneous distributed systems.
\newblock In {\em Advances in Neural Information Processing Systems}, 2015.

\bibitem{choi2020starganv2}
Yunjey Choi, Youngjung Uh, Jaejun Yoo, and Jung-Woo Ha.
\newblock Stargan v2: Diverse image synthesis for multiple domains.
\newblock In {\em Proceedings of the IEEE Conference on Computer Vision and
  Pattern Recognition}, 2020.

\bibitem{chollet2015keras}
Fran{\c{c}}ois Chollet et~al.
\newblock keras, 2015.

\bibitem{chong2019effectively}
Min~Jin Chong and David Forsyth.
\newblock Effectively unbiased fid and inception score and where to find them.
\newblock In {\em CVPR}, 2020.

\bibitem{clark2015pillow}
Alex Clark.
\newblock Pillow (pil fork) documentation, 2015.

\bibitem{deng2009imagenet}
Jia Deng, Wei Dong, Richard Socher, Li-Jia Li, Kai Li, and Li Fei-Fei.
\newblock Imagenet: A large-scale hierarchical image database.
\newblock In {\em IEEE Conference on Computer Vision and Pattern Recognition
  (CVPR)}, 2009.

\bibitem{denton2015deep}
Emily~L Denton, Soumith Chintala, Rob Fergus, et~al.
\newblock Deep generative image models using a laplacian pyramid of adversarial
  networks.
\newblock In {\em Advances in Neural Information Processing Systems}, 2015.

\bibitem{dinh2017density}
Laurent Dinh, Jascha Sohl-Dickstein, and Samy Bengio.
\newblock Density estimation using real nvp.
\newblock In {\em International Conference on Learning Representations (ICLR)},
  2017.

\bibitem{donahue2019large}
Jeff Donahue and Karen Simonyan.
\newblock Large scale adversarial representation learning.
\newblock In {\em Advances in Neural Information Processing Systems}, 2019.

\bibitem{dosovitskiy2016generating}
Alexey Dosovitskiy and Thomas Brox.
\newblock Generating images with perceptual similarity metrics based on deep
  networks.
\newblock In {\em Advances in Neural Information Processing Systems}, 2016.

\bibitem{du2019implicit}
Yilun Du and Igor Mordatch.
\newblock Implicit generation and generalization in energy-based models.
\newblock In {\em Advances in Neural Information Processing Systems}, 2019.

\bibitem{engstrom2019exploring}
Logan Engstrom, Brandon Tran, Dimitris Tsipras, Ludwig Schmidt, and Aleksander
  Madry.
\newblock Exploring the landscape of spatial robustness.
\newblock In {\em International Conference on Machine Learning}, pages
  1802--1811. PMLR, 2019.

\bibitem{foley1996computer}
James~D Foley, Foley~Dan Van, Andries Van~Dam, Steven~K Feiner, John~F Hughes,
  and J Hughes.
\newblock {\em Computer graphics: principles and practice}, volume 12110.
\newblock Addison-Wesley Professional, 1996.

\bibitem{foley1994introduction}
James~D Foley, Andries Van~Dam, Steven~K Feiner, John~F Hughes, and Richard~L
  Phillips.
\newblock {\em Introduction to computer graphics}, volume~55.
\newblock Addison-Wesley Reading, 1994.

\bibitem{forsyth2012computer}
David~A Forsyth and Jean Ponce.
\newblock {\em Computer vision: a modern approach}.
\newblock Pearson,, 2012.

\bibitem{gatys2015neural}
Leon~A Gatys, Alexander~S Ecker, and Matthias Bethge.
\newblock Image style transfer using convolutional neural networks.
\newblock In {\em IEEE Conference on Computer Vision and Pattern Recognition
  (CVPR)}, 2016.

\bibitem{gonzalez2002digital}
Rafael~C Gonzalez, Richard~E Woods, et~al.
\newblock Digital image processing, 2002.

\bibitem{goodfellow2014generative}
Ian Goodfellow, Jean Pouget-Abadie, Mehdi Mirza, Bing Xu, David Warde-Farley,
  Sherjil Ozair, Aaron Courville, and Yoshua Bengio.
\newblock Generative adversarial nets.
\newblock In {\em Advances in Neural Information Processing Systems}, 2014.

\bibitem{hendrycks2019benchmarking}
Dan Hendrycks and Thomas Dietterich.
\newblock Benchmarking neural network robustness to common corruptions and
  perturbations.
\newblock {\em ICLR}, 2019.

\bibitem{heusel2017gans}
Martin Heusel, Hubert Ramsauer, Thomas Unterthiner, Bernhard Nessler, and Sepp
  Hochreiter.
\newblock {GANs} trained by a two time-scale update rule converge to a local
  {Nash} equilibrium.
\newblock In {\em Advances in Neural Information Processing Systems}, 2017.

\bibitem{howard2017mobilenets}
Andrew~G Howard, Menglong Zhu, Bo Chen, Dmitry Kalenichenko, Weijun Wang,
  Tobias Weyand, Marco Andreetto, and Hartwig Adam.
\newblock Mobilenets: Efficient convolutional neural networks for mobile vision
  applications.
\newblock {\em arXiv preprint arXiv:1704.04861}, 2017.

\bibitem{huang2018multimodal}
Xun Huang, Ming-Yu Liu, Serge Belongie, and Jan Kautz.
\newblock Multimodal unsupervised image-to-image translation.
\newblock {\em European Conference on Computer Vision (ECCV)}, 2018.

\bibitem{johnson2016perceptual}
Justin Johnson, Alexandre Alahi, and Li Fei-Fei.
\newblock Perceptual losses for real-time style transfer and super-resolution.
\newblock In {\em European Conference on Computer Vision (ECCV)}, 2016.

\bibitem{karras2018progressive}
Tero Karras, Timo Aila, Samuli Laine, and Jaakko Lehtinen.
\newblock Progressive growing of gans for improved quality, stability, and
  variation.
\newblock In {\em International Conference on Learning Representations (ICLR)},
  2018.

\bibitem{karras2020training}
Tero Karras, Miika Aittala, Janne Hellsten, Samuli Laine, Jaakko Lehtinen, and
  Timo Aila.
\newblock Training generative adversarial networks with limited data.
\newblock {\em NIPS}, 33, 2020.

\bibitem{karras2021alias}
Tero Karras, Miika Aittala, Samuli Laine, Erik H\"ark\"onen, Janne Hellsten,
  Jaakko Lehtinen, and Timo Aila.
\newblock Alias-free generative adversarial networks.
\newblock In {\em Proc. NeurIPS}, 2021.

\bibitem{karras2019style}
Tero Karras, Samuli Laine, and Timo Aila.
\newblock A style-based generator architecture for generative adversarial
  networks.
\newblock In {\em IEEE Conference on Computer Vision and Pattern Recognition
  (CVPR)}, 2019.

\bibitem{karras2020analyzing}
Tero Karras, Samuli Laine, Miika Aittala, Janne Hellsten, Jaakko Lehtinen, and
  Timo Aila.
\newblock Analyzing and improving the image quality of stylegan.
\newblock {\em IEEE Conference on Computer Vision and Pattern Recognition
  (CVPR)}, 2020.

\bibitem{keys81cubic}
R. Keys.
\newblock Cubic convolution interpolation for digital image processing.
\newblock {\em IEEE Transactions on Acoustics, Speech, and Signal Processing},
  29(6):1153--1160, 1981.

\bibitem{kingma2018glow}
Diederik~P Kingma and Prafulla Dhariwal.
\newblock Glow: Generative flow with invertible 1x1 convolutions.
\newblock In {\em Advances in Neural Information Processing Systems}, 2018.

\bibitem{kingma2014auto}
Diederik~P Kingma and Max Welling.
\newblock Auto-encoding variational bayes.
\newblock {\em International Conference on Learning Representations (ICLR)},
  2014.

\bibitem{kynkaanniemi2019improved}
Tuomas Kynk{\"a}{\"a}nniemi, Tero Karras, Samuli Laine, Jaakko Lehtinen, and
  Timo Aila.
\newblock Improved precision and recall metric for assessing generative models.
\newblock In {\em Advances in Neural Information Processing Systems}, 2019.

\bibitem{lecun1998gradient}
Yann LeCun, L{\'e}on Bottou, Yoshua Bengio, and Patrick Haffner.
\newblock Gradient-based learning applied to document recognition.
\newblock {\em Proceedings of the IEEE}, 86(11):2278--2324, 1998.

\bibitem{lehmann2006testing}
Erich~L Lehmann and Joseph~P Romano.
\newblock {\em Testing statistical hypotheses}.
\newblock Springer Science \& Business Media, 2006.

\bibitem{liu2019few}
Ming-Yu Liu, Xun Huang, Arun Mallya, Tero Karras, Timo Aila, Jaakko Lehtinen,
  and Jan Kautz.
\newblock Few-shot unsupervised image-to-image translation.
\newblock In {\em IEEE International Conference on Computer Vision (ICCV)},
  2019.

\bibitem{liu2020diverse}
Steven Liu, Tongzhou Wang, David Bau, Jun-Yan Zhu, and Antonio Torralba.
\newblock Diverse image generation via self-conditioned gans.
\newblock In {\em IEEE Conference on Computer Vision and Pattern Recognition
  (CVPR)}, 2020.

\bibitem{lopez2016revisiting}
David Lopez-Paz and Maxime Oquab.
\newblock Revisiting classifier two-sample tests.
\newblock In {\em ICLR}, 2017.

\bibitem{lucic2017gans}
Mario Lucic, Karol Kurach, Marcin Michalski, Sylvain Gelly, and Olivier
  Bousquet.
\newblock Are gans created equal? a large-scale study.
\newblock In {\em Advances in Neural Information Processing Systems}, 2018.

\bibitem{mairal2014convolutional}
Julien Mairal, Piotr Koniusz, Zaid Harchaoui, and Cordelia Schmid.
\newblock Convolutional kernel networks.
\newblock {\em Advances in neural information processing systems},
  27:2627--2635, 2014.

\bibitem{nijkamp2020anatomy}
Erik Nijkamp, Mitch Hill, Tian Han, Song-Chun Zhu, and Ying~Nian Wu.
\newblock On the anatomy of mcmc-based maximum likelihood learning of
  energy-based models.
\newblock In {\em AAAI Conference on Artificial Intelligence (AAAI)}, 2020.

\bibitem{nyquist1928certain}
Harry Nyquist.
\newblock Certain topics in telegraph transmission theory.
\newblock {\em Transactions of the American Institute of Electrical Engineers},
  47(2):617--644, 1928.

\bibitem{oord2016conditional}
Aaron van~den Oord, Nal Kalchbrenner, Oriol Vinyals, Lasse Espeholt, Alex
  Graves, and Koray Kavukcuoglu.
\newblock Conditional image generation with pixelcnn decoders.
\newblock In {\em Advances in Neural Information Processing Systems}, 2016.

\bibitem{oppenheim1999discrete}
Alan~V. Oppenheim, Ronald~W. Schafer, and John~R. Buck.
\newblock {\em Discrete-Time Signal Processing}.
\newblock Pearson, 2nd edition, 1999.

\bibitem{park2019SPADE}
Taesung Park, Ming-Yu Liu, Ting-Chun Wang, and Jun-Yan Zhu.
\newblock Semantic image synthesis with spatially-adaptive normalization.
\newblock In {\em IEEE Conference on Computer Vision and Pattern Recognition
  (CVPR)}, 2019.

\bibitem{dcvae21}
Gaurav Parmar, Dacheng Li, Kwonjoon Lee, and Zhuowen Tu.
\newblock Dual contradistinctive generative autoencoder.
\newblock In {\em IEEE Conference on Computer Vision and Pattern Recognition
  (CVPR)}, 2021.

\bibitem{paszke2019pytorch}
Adam Paszke, Sam Gross, Francisco Massa, Adam Lerer, James Bradbury, Gregory
  Chanan, Trevor Killeen, Zeming Lin, Natalia Gimelshein, Luca Antiga, et~al.
\newblock Pytorch: An imperative style, high-performance deep learning library.
\newblock In {\em Advances in Neural Information Processing Systems}, 2019.

\bibitem{qian2021blending}
Shengju Qian, Hao Shao, Yi Zhu, Mu Li, and Jiaya Jia.
\newblock Blending anti-aliasing into vision transformer.
\newblock In {\em Thirty-Fifth Conference on Neural Information Processing
  Systems}, 2021.

\bibitem{quiring2020adversarial}
Erwin Quiring, David Klein, Daniel Arp, Martin Johns, and Konrad Rieck.
\newblock Adversarial preprocessing: Understanding and preventing image-scaling
  attacks in machine learning.
\newblock In {\em 29th $\{$USENIX$\}$ Security Symposium ($\{$USENIX$\}$
  Security 20)}, pages 1363--1380, 2020.

\bibitem{radford2015unsupervised}
Alec Radford, Luke Metz, and Soumith Chintala.
\newblock Unsupervised representation learning with deep convolutional
  generative adversarial networks.
\newblock In {\em International Conference on Learning Representations (ICLR)},
  2016.

\bibitem{ravuri2019classification}
Suman Ravuri and Oriol Vinyals.
\newblock Classification accuracy score for conditional generative models.
\newblock In {\em Advances in Neural Information Processing Systems}, 2019.

\bibitem{razavi2019generating}
Ali Razavi, Aaron van~den Oord, and Oriol Vinyals.
\newblock Generating diverse high-fidelity images with vq-vae-2.
\newblock In {\em NIPS}, 2019.

\bibitem{sajjadi2018assessing}
Mehdi~SM Sajjadi, Olivier Bachem, Mario Lucic, Olivier Bousquet, and Sylvain
  Gelly.
\newblock Assessing generative models via precision and recall.
\newblock In {\em Advances in Neural Information Processing Systems}, 2018.

\bibitem{salakhutdinov2009deep}
Ruslan Salakhutdinov and Geoffrey Hinton.
\newblock Deep boltzmann machines.
\newblock In {\em Artificial intelligence and statistics}, pages 448--455,
  2009.

\bibitem{salimans2016improved}
Tim Salimans, Ian Goodfellow, Wojciech Zaremba, Vicki Cheung, Alec Radford, and
  Xi Chen.
\newblock Improved techniques for training gans.
\newblock In {\em Advances in Neural Information Processing Systems}, 2016.

\bibitem{scherer2010evaluation}
Dominik Scherer, Andreas M{\"u}ller, and Sven Behnke.
\newblock Evaluation of pooling operations in convolutional architectures for
  object recognition.
\newblock In {\em International conference on artificial neural networks},
  pages 92--101. Springer, 2010.

\bibitem{Seitzer2020FID}
Maximilian Seitzer.
\newblock {pytorch-fid: FID Score for PyTorch}.
\newblock \url{https://github.com/mseitzer/pytorch-fid}, August 2020.
\newblock Version 0.1.1.

\bibitem{shrivastava2017learning}
Ashish Shrivastava, Tomas Pfister, Oncel Tuzel, Josh Susskind, Wenda Wang, and
  Russ Webb.
\newblock Learning from simulated and unsupervised images through adversarial
  training.
\newblock In {\em IEEE Conference on Computer Vision and Pattern Recognition
  (CVPR)}, 2017.

\bibitem{sifre2014rigid}
Laurent Sifre and St{\'e}phane Mallat.
\newblock Rigid-motion scattering for texture classification.
\newblock {\em arXiv preprint arXiv:1403.1687}, 2014.

\bibitem{szegedy2015rethinking}
Christian Szegedy, Vincent Vanhoucke, Sergey Ioffe, Jon Shlens, and Zbigniew
  Wojna.
\newblock Rethinking the inception architecture for computer vision.
\newblock In {\em IEEE Conference on Computer Vision and Pattern Recognition
  (CVPR)}, 2016.

\bibitem{szeliski2010computer}
Richard Szeliski.
\newblock {\em Computer vision: algorithms and applications}.
\newblock Springer Science \& Business Media, 2010.

\bibitem{taori2020measuring}
Rohan Taori, Achal Dave, Vaishaal Shankar, Nicholas Carlini, Benjamin Recht,
  and Ludwig Schmidt.
\newblock Measuring robustness to natural distribution shifts in image
  classification.
\newblock In {\em Advances in Neural Information Processing Systems (NeurIPS)},
  2020.

\bibitem{theis2015note}
Lucas Theis, A{\"a}ron van~den Oord, and Matthias Bethge.
\newblock A note on the evaluation of generative models.
\newblock In {\em ICLR}, 2016.

\bibitem{wallace1992jpeg}
Gregory~K Wallace.
\newblock The jpeg still picture compression standard.
\newblock {\em IEEE transactions on consumer electronics}, 38(1):xviii--xxxiv,
  1992.

\bibitem{wang2020cnn}
Sheng-Yu Wang, Oliver Wang, Richard Zhang, Andrew Owens, and Alexei~A. Efros.
\newblock Cnn-generated images are surprisingly easy to spot... for now.
\newblock In {\em IEEE Conference on Computer Vision and Pattern Recognition
  (CVPR)}, 2020.

\bibitem{xiao2019seeing}
Qixue Xiao, Yufei Chen, Chao Shen, Yu Chen, and Kang Li.
\newblock Seeing is not believing: Camouflage attacks on image scaling
  algorithms.
\newblock In {\em 28th $\{$USENIX$\}$ Security Symposium ($\{$USENIX$\}$
  Security 19)}, pages 443--460, 2019.

\bibitem{yu15lsun}
Fisher Yu, Yinda Zhang, Shuran Song, Ari Seff, and Jianxiong Xiao.
\newblock Lsun: Construction of a large-scale image dataset using deep learning
  with humans in the loop.
\newblock {\em arXiv preprint arXiv:1506.03365}, 2015.

\bibitem{zhang2019consistency}
Han Zhang, Zizhao Zhang, Augustus Odena, and Honglak Lee.
\newblock Consistency regularization for generative adversarial networks.
\newblock In {\em International Conference on Learning Representations (ICLR)},
  2020.

\bibitem{zhang2020consistency}
Han Zhang, Zizhao Zhang, Augustus Odena, and Honglak Lee.
\newblock Consistency regularization for generative adversarial networks.
\newblock In {\em International Conference on Learning Representations (ICLR)},
  2020.

\bibitem{zhang2019making}
Richard Zhang.
\newblock Making convolutional networks shift-invariant again.
\newblock In {\em International Conference on Machine Learning (ICML)}, 2019.

\bibitem{zhang2018unreasonable}
Richard Zhang, Phillip Isola, Alexei~A Efros, Eli Shechtman, and Oliver Wang.
\newblock The unreasonable effectiveness of deep features as a perceptual
  metric.
\newblock In {\em IEEE Conference on Computer Vision and Pattern Recognition
  (CVPR)}, 2018.

\bibitem{zhao2020diffaugment}
Shengyu Zhao, Zhijian Liu, Ji Lin, Jun-Yan Zhu, and Song Han.
\newblock Differentiable augmentation for data-efficient gan training.
\newblock In {\em Advances in Neural Information Processing Systems (NeurIPS)},
  2020.

\bibitem{zhou2019hype}
Sharon Zhou, Mitchell~L Gordon, Ranjay Krishna, Austin Narcomey, Li Fei-Fei,
  and Michael~S Bernstein.
\newblock Hype: A benchmark for human eye perceptual evaluation of generative
  models.
\newblock In {\em Advances in Neural Information Processing Systems}, 2019.

\bibitem{zhu2016generative}
Jun-Yan Zhu, Philipp Kr{\"a}henb{\"u}hl, Eli Shechtman, and Alexei~A Efros.
\newblock Generative visual manipulation on the natural image manifold.
\newblock In {\em European Conference on Computer Vision (ECCV)}, 2016.

\bibitem{zhu2017unpaired}
Jun-Yan Zhu, Taesung Park, Phillip Isola, and Alexei~A Efros.
\newblock Unpaired image-to-image translation using cycle-consistent
  adversarial networks.
\newblock In {\em IEEE International Conference on Computer Vision (ICCV)},
  2017.

\bibitem{zou2020delving}
Xueyan Zou, Fanyi Xiao, Zhiding Yu, and Yong~Jae Lee.
\newblock Delving deeper into anti-aliasing in convnets.
\newblock In {\em BMVC}, 2020.

\end{thebibliography}
}

\clearpage
\section{Appendix}

\subsection{Interpolation Filters}
\lblsec{ap_filters}
\reffig{resizing_steps} shows the image downsampling procedure. When the resizing ratio is an integer, downsampling can be implemented as a discrete convolution with an interpolation kernel, followed by subsampling. As discussed in \refsec{image_resizing}, the kernel needs to be widened according to the resizing ratio to prevent aliasing in the resized image.
All of the commonly used interpolation filters are separable, meaning the two-dimensional interpolation $K(x,y)$ over an image can be decomposed into one-dimensional interpolations $u(s)$ along each dimension. Here, $x, y$ represent the spatial coordinates for the two-dimensional case and $s$ represents the spatial coordinates for the one-dimension case. 
$$ K(x,y) = u(x)~u(y)$$
Next we describe each of the different interpolation functions in one-dimension. 

\myparagraph{Nearest Neighbor Interpolation.}
The simplest form of image interpolation is the nearest neighbor interpolation which only considers the value of the neighboring point. This is equivalent to interpolating with the function shown below. 
\begin{equation}
    u(s) = \begin{cases}
      1 & \abs{s} < 0.5 \\
      0 & \text{otherwise}
    \end{cases} 
\vspace{-2pt}
\lbleq{kernel_nn}
\end{equation}

\myparagraph{Bilinear Interpolation.}
The bilinear image interpolation corresponds to interpolating using the triangle filter defined below.
\begin{equation}
    u(s) = max(1-\abs{s}, 0)
\vspace{-2pt}
\lbleq{kernel_bilinear}
\end{equation}

\myparagraph{Lanczos Interpolation.}
The Lanczos image interpolation is the normalized sinc functions windowed by the Lanczos window $w(s)$.
\begin{equation}
    u(s) = w(s)\text{sinc}(s)
\vspace{-2pt}
\lbleq{kernel_lanczos}
\end{equation}
\begin{equation}
    w(s) = \begin{cases}
      \text{sinc}(s/n) & \abs{s} < n \\
      0 & \text{otherwise}
    \end{cases} 
\vspace{-2pt}
\lbleq{window}
\end{equation}
$n$ is typically 2 or 3. 

\myparagraph{Bicubic Interpolation.} The bicubic interpolation \cite{keys81cubic} uses the interpolation kernel $u(s)$. 
\begin{equation}
    u(s) = \begin{cases}
      (\alpha + 2) \abs{s}^3 - (\alpha + 3)\abs{s}^2 + 1 & \abs{s} < 1 \\
      \alpha\abs{s}^3 - 5\alpha\abs{s}^2 + 8\alpha\abs{s} - 4\alpha & 1 < \abs{s} < 2 \\
      0 & \abs{s} > 2
    \end{cases} 
\vspace{-2pt}
\lbleq{kernel_bicubic}
\end{equation}
The common choices for the free parameter $\alpha$ are $-0.5, -0.75, -1.0$.

\begin{figure}[t!]
    \centering
    \includegraphics[width=1.0\linewidth]{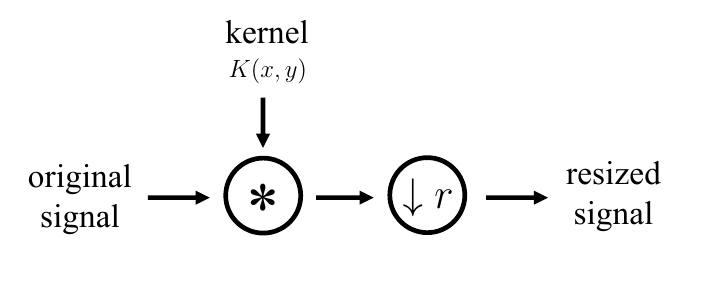}
  \vspace{-8mm}
  \caption{
  \textbf{Steps in Resizing.} We illustrate the downsampling procedure for integer resizing factors. First the input original signal is discretely convolved with the interpolation kernel $K(x,y)$. In order to antialias or prefilter the input signal, the interpolation kernel \textit{should} be stretched, according to the downsampling ratio r. Next, the convolved signal is subsampled to obtain the final resized signal. 
}
  \lblfig{resizing_steps}
  \vspace{-5mm}
\end{figure}

\myparagraph{Filter scaling.} As shown in~\reffig{kernels} and discussed in~\refsec{image_resizing}, whether to adapt the kernel width to the downsampling factor has a large qualitative and quantitative effect on the downsampled image. The continuous filter $u(s)$ is sampled at a set of discrete locations and yield a discrete filter and normalized to sum to $1$. The difference between adaptive and non-adaptive filters arise at which locations are sampled.

For an adaptive filter, $u(s)$ is sampled at $s\in \{..., -\frac{3}{2r}, -\frac{1}{2r}, \frac{1}{2r}, \frac{3}{2r}, ...\}$ for even downsampling factors and $s\in \{..., -\frac{2}{r}, -\frac{1}{r}, 0, \frac{1}{r}, \frac{2}{r}, ... \}$ for odd factors. The filter width widens with larger downsampling factor $r$.

For a non-adaptive filter, $s\in \{..., -\frac{3}{2}, -\frac{1}{2}, \frac{1}{2}, \frac{3}{2}, ...\}$ for even factors and $s\in \{..., -2, -1, 0, 1, 2, ...\}$ for odd factors. Notice the sampling locations do not scale as a function of downsampling factor $r$.

From here, one can observe why a non-adaptive filter behaves similarly to nearest. For even factors, plugging in the sampling locations yields a 2-tap filter $\{..., 0, \frac{1}{2}, \frac{1}{2}, 0, ...\}$. For odd factors, yields delta function $\{..., 0, 0, 1, 0, 0, ...\}$ for all filters. In contrast, for an adaptive filter, a $r=2$ bilinear downsample yields a 4-tap $\{\frac{1}{8}, \frac{3}{8}, \frac{3}{8}, \frac{1}{8}\}$ filter, $r=4$ yields an 8-tap filter, etc.

\subsection{JPEG Compression.}
In Sections \ref{sec:image_format} and \ref{sec:resize_fid_jpeg} in the main paper, we discuss the compression of images and the effects on evaluation metrics such as FID and KID. Next, we detail the JPEG compression protocol in \reffig{jpeg_steps}, and outline the three steps that result in a loss of information. Motivated by the observation that the human vision is less sensitive to color components, the first lossy step is the subsampling of color channels Cr, Cb after the color space transformation. Next, the image channels are divided into smaller $8\times8$ blocks and the Discrete Cosine Transformation (DCT) is computed. The DCT coefficients are subsequently divided by the quantization table to suppress the higher frequencies and rounded to integers. The quantization table is determined by the user specified "quality" option (0-100) and controls the tradeoff between the storage space and image information retained. 
When the quality option is set to 100, the color subsampling and integer rounding are the primary sources of information loss. 

\begin{figure*}[t]
    \centering
    \includegraphics[width=1.0\linewidth]{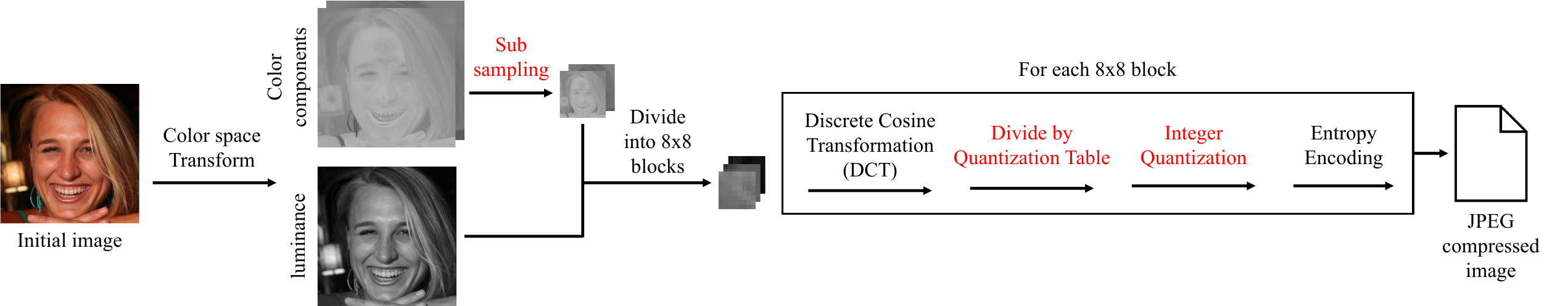}
  \vspace{-3mm}
  \caption{
  \textbf{Steps in JPEG Compression.} We illustrate the different steps involved in JPEG compression of images. First, the initial image is converted from the RGB color space to the YCrCb space. Next, the luminance channel and the subsampled color components are divided into $8\times8$ blocks. The DCT coefficients of each block are subsequently quantized and encoded. The lossy steps of the process are highlighted in red. }
  \lblfig{jpeg_steps}
  \vspace{-5mm}
\end{figure*}

\subsection{Library Implementation Details.}
The library implementations used for the comparisons are detailed below. 
\begin{itemize}
  \setlength{\itemsep}{1pt}
  \setlength{\parskip}{0pt}
  \setlength{\parsep}{0pt}
    \item \textbf{Pillow Image Library (PIL) v8.0.1~\cite{clark2015pillow}}: We use the standard \texttt{Image.resize} function; the library provides consistently antialiased results across filters.
    \item \textbf{OpenCV v4.5.5~\cite{bradski2000opencv}:} We use the standard \texttt{cv2.resize} function.
    \item \textbf{TensorFlow (TF) v2.0~\cite{abadi2016tensorflow}: } For the comparisons in this section we use the flags used by the original TensorFlow implementation of FID. The TensorFlow library has changed substantially through the versions. In this work we use the new TensorFlow version 2.0. Note that the newer version of the library has an optional flag \texttt{antialias}. However this option is set to \texttt{False} by default and not used in the current FID implementations. 
    \item \textbf{PyTorch v1.9~\cite{paszke2019pytorch}: }
    We use the differentiable function \texttt{F.interpolate} on data tensors.\footnote{A separate function, \texttt{torchvision.transforms.Resize}, is a wrapper around the PIL library and is often used in the data pre-processing step.}
    This resizing method has been used by popular PyTorch implementations of FID \cite{Seitzer2020FID}. 
    \item \textbf{MXNet v1.8~\cite{chen2015mxnet}: } The resizing method provided in the MXNet framework is a wrapper around the OpenCV \cite{bradski2000opencv} implementation. 
    \item \textbf{Keras v2.6.0~\cite{chollet2015keras}: } The library is built on top of the TensorFlow ~\cite{abadi2016tensorflow} framework and shares the implementation for resizing images. 
\end{itemize}

\subsection{Additional resizing example}
In \reffig{teaser} in the main paper, we showed an example resizing a sparse circle. We observe that when the bicubic, lanczos, and bilinear filters do not adjust their filter widths to the downsampling factor, aliasing patterns occur. This occurs in several libraries, including the settings used in PyTorch and TensorFlow for FID calculation.

Here, in \reffig{downsampling_ext}, we show an image \textit{with varying frequency content}, in order to further illustrate the behavior of different downsampling filters and implementations. The input is of size $200$ and is downsampled by $5\times$ to resolution $400$. The input image is of concentric circles, with low frequency in the middle and increasing frequency towards the outside.

When the image is heavily downsampled, the high frequencies on the outside cannot be represented by a low resolution. As seen in the bottom left of \reffig{downsampling_ext}, naive subsampling results in heavy aliasing, with a grid of additional circles being hallucinated in the output. A well-filtered downsampling result would instead retain the circle in the middle, while filtering out the high-frequency content into gray. This is observed in implementations where the filter is adjusted based on the downsampling factor -- namely the PIL implementations of bicubic, lanczos, and bilinear and Tensorflow with antialias flag set as \textsc{True}. As before, using a fixed-width filter, as in the other rows, results in heavy aliasing.

In addition, we also show the \textit{area} filter. Here, we observe a mixed results. Because implementations of the area filter do adjust to the downsampling factor across all libraries, the aliasing is not as apparent as in naive subsampling, or the fixed-width implementations of bicubic, lanczos, and bilinear. However, as described in L417 in the main paper, this particular filter corresponds to a \textit{box}, or rectangular filter, which does not have strong antialiasing properties as the other filters. As a result, there are significantly more artifacts (additional hallucinated concentric circles) compared to the stronger filters (bicubic, lanczos, and bilinear) which adjust the filter widths.

In conclusion, this shows that in practical implementations, the variations in whether the filter width \textit{and} the actual filter type both have an effect on the aliasing artifacts on the output.

\begin{figure*}[t]
    \centering
    \includegraphics[width=1.0\linewidth]{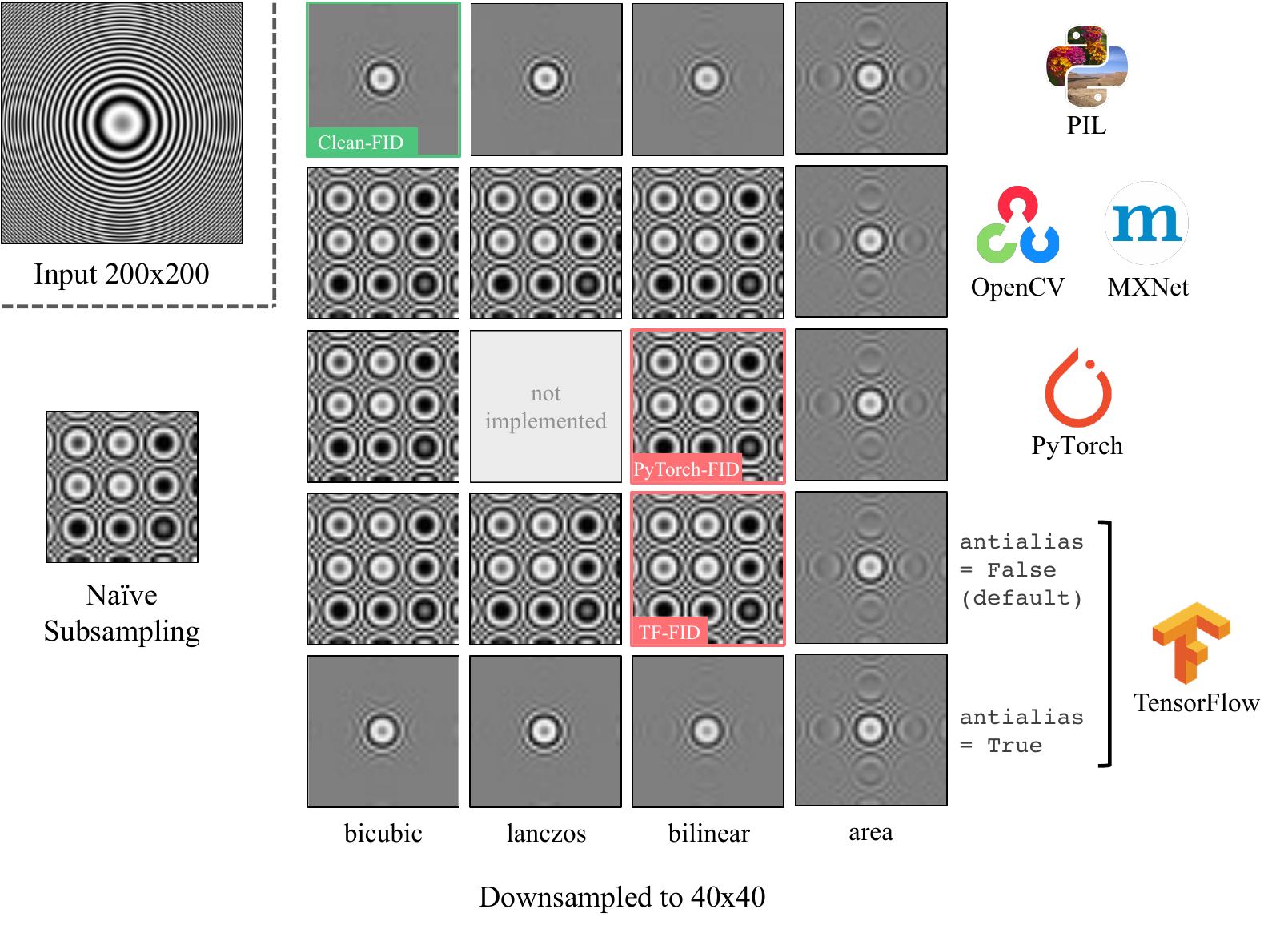}
  \vspace{-8mm}
  \caption{
  \textbf{Downsampling an image.} We downsample an image containing multiple frequencies from an input size of $200 \times 200$ to $40 \times 40$. (We encourage viewing this figure without zooming-out on a digital display.)}
  \lblfig{downsampling_ext}
  \vspace{-5mm}
\end{figure*}

\end{document}